%% file: main.tex
\documentclass{article}
\input{preamble/preamble}

\input{preamble/preamble_acronyms}
\input{preamble/preamble_math}

\input{preamble/preamble_tikz}

\usepackage{rotating}
\usepackage{multirow}

\usepackage[accepted]{icml2018}

\icmltitlerunning{Cycle-Consistent Adversarial Learning as 
				          Approximate Bayesian Inference}

\begin{document}

\twocolumn[
\icmltitle{Cycle-Consistent Adversarial Learning as \\ 
		       Approximate Bayesian Inference}

\icmlsetsymbol{equal}{*}

\begin{icmlauthorlist}
\icmlauthor{Louis C.~Tiao}{usyd}
\icmlauthor{Edwin V.~Bonilla}{unsw}
\icmlauthor{Fabio Ramos}{usyd}
\end{icmlauthorlist}

\icmlaffiliation{usyd}{University of Sydney, Sydney, Australia}
\icmlaffiliation{unsw}{University of New South Wales, Sydney, Australia}

\icmlcorrespondingauthor{Louis Tiao}{louis.tiao@sydney.edu.au}

\icmlkeywords{Machine Learning, ICML}

\vskip 0.3in
]



\printAffiliationsAndNotice{}  

\begin{abstract}
We formalize the problem of learning inter-domain correspondences in 
the absence of paired data as Bayesian inference in a \gls{LVM}, where 
one seeks the underlying hidden representations of entities from one 
domain as entities from the other domain.
First, we introduce \emph{\acrlongpl{ILVM}}, where the prior over 
hidden representations can be specified flexibly as an \emph{implicit} 
distribution.
Next, we develop a new \gls{VI} algorithm for this model based on 
minimization of the \emph{symmetric} \gls{KL} divergence between a 
variational joint and the exact joint distribution.
Lastly, we demonstrate that the state-of-the-art \gls{CYCLEGAN} models
can be derived as a special case within our proposed \gls{VI} framework,
thus establishing its connection to approximate Bayesian inference methods.
\end{abstract}

\section{Introduction}
\label{sec:introduction}

Learning correspondences between entities from different domains is 
an important and challenging problem in machine learning, especially 
in the \emph{absence of paired data}. Consider for example the task of 
image-to-image translation where we want to learn a mapping from an 
image in a source domain, such as a photograph of a natural scene, to 
a corresponding image in a target domain, such as the realization of 
such a scene in an 1860s celebrated artist's signature impressionistic 
style. The shortage of ground-truth pairings from the source domain to 
the target domain renders standard supervised approaches infeasible,
thus motivating the need for unsupervised learning approaches.

Within these unsupervised approaches, a number of recently proposed 
\acrfull{CYCLEGAN} methods have achieved remarkable success in 
addressing this problem \citep{pmlr-v70-kim17a,Zhu2017}. 
As their name suggests, these approaches are based upon two heuristics: 
(i) adversarial learning and (ii) cycle consistency. 
The former, adversarial learning \citep{NIPS2014_5423}, allows images 
in the source domain to be translated to output images that, to an 
auxiliary discriminator, are indistinguishable from images in the 
target domain, thereby matching their distributions.
However, while distribution matching is necessary, it is insufficient 
to guarantee one-to-one mappings between the images, as the problem is 
heavily under-constrained. 
Briefly stated, the cycle-consistency is the constraint that an image 
mapped to a target domain should be \emph{representable} in the 
original domain. It is this constraint that significantly shrinks the 
space of possible solutions.
 
Beyond the empirical risk minimization framework motivated intuitively by the two conceptually simple principles mentioned above, the original \gls{CYCLEGAN} formulation lacks any further theoretical justification. This hinders (i) understanding  of the distributional assumptions of all the variables of interest;  (ii) how these are combined in terms of prior knowledge and observational assumptions; and  (iii) whether more general methods can be used for estimating their parameters. 
In contrast, \glspl{LVM} offer a principled framework for probabilistic reasoning about random variables, their statistical properties and dependency structures, with the goal of capturing the underlying data-generation process. Besides providing sound quantification of uncertainty, a \acrshort{LVM} allows us to disentangle our  modeling assumptions from the inference machinery we use to reason about the variables in the model. Interpreting standard methods from a Bayesian perspective has  contributed significantly to the understanding of these methods and to the development of new approaches. Examples of this include the seminal work on \gls{PPCA} \citep{Tipping1999} and more recently,
the advances in understanding Dropout \citep{Gal2015}. 
Our contributions are threefold, and are summarized below.

\parhead{First contribution} (\cref{sec:implicit_latent_variable_models}). We formulate the problem of learning correspondences between domains using unpaired data from a \gls{LVM} perspective, where entities in one domain are latent representations of  entities in the other domain. 
Crucial to our approach is to consider an \emph{implicit} prior over these hidden representations,
i.e.~a prior that is not given by a prescribed distribution such as a Gaussian, but instead, is provided via samples from an unknown process. 

\parhead{Second contribution} (\cref{sec:variational_inference_with_implicit_priors,sec:symmetric_joint_matching_variational_inference}). We develop a new scalable variational inference algorithm for these types of models. Unlike standard  \gls{VI} approaches that approximate the posterior over the latent variables, we approximate the exact joint distribution over latent and observed variables with a variational joint. Furthermore,  we carry out this approximation via the minimization of the symmetric  \gls{KL} divergence between these joints. This is in stark contrast with traditional \gls{VI} approaches that minimize the forward \gls{KL} divergence, as the symmetric \gls{KL} divergence between the posteriors is intractable. 

\parhead{Third contribution} (\cref{sec:deriving_cyclegan_as_a_special_case}). Finally, we show that \glspl{CYCLEGAN}, as proposed by \cite{Zhu2017,pmlr-v70-kim17a} independently, can be recovered by applying  our approximate inference algorithm in a \gls{LVM} for domain correspondence  with unpaired data. In this model, the prior is specified by an implicit \emph{empirical} distribution and the observed variables are generated by a nonlinear function of its underlying latent variable.  Intuitively, while using the symmetric \gls{KL} divergence between the approximate and true joint distributions yields \gls{GAN}-like objectives between the domains, different specifications of the likelihood and the approximate posterior yield the cycle-consistency part of the loss in \glspl{CYCLEGAN}.  





\section{Implicit Latent Variable Models}
\label{sec:implicit_latent_variable_models}

\Acrfullpl{LVM} are an indispensable tool for uncovering the hidden 
representations of observed data. 
In a \gls{LVM}, an observation $\mbx$ is assumed governed by its 
underlying hidden variable $\mbz$, which is drawn from a prior 
$p(\mbz)$ and related to $\mbx$ through the likelihood 
$p_{\mbtheta}(\mbx \g \mbz)$. 
Accordingly, the joint density of $\mbx$ and $\mbz$ is given by
\begin{equation} \label{eq:joint}
  p_{\mbtheta}(\mbx,\mbz) = p_{\mbtheta}(\mbx\g\mbz) p(\mbz).
\end{equation}
Given data distribution $q^*(\mbx)$ and a finite collection 
$\mbX = \{ \mbx_n \}_{n=1}^N$ of observations $\mbx_n \sim q^*(\mbx)$, 
and the set of corresponding latent variables 
$\mbZ = \{ \mbz_n \}_{n=1}^N$, the joint over all variables factorizes 
as, 
\begin{equation}
  p_{\mbtheta}(\mbX,\mbZ) 
  = 
  \prod_{n=1}^N p_{\mbtheta}(\mbx_n,\mbz_n).
\end{equation}
The graphical representation of this model is depicted in 
\cref{fig:graphical_representation}.

\subsection{Prescribed Likelihood}
\label{sub:likelihood}

We specify the likelihood through a mapping $\cF_{\mbtheta}$ that 
takes as input random noise $\mbxi$ and latent variable $\mbz$,
\begin{equation} \label{eq:likelihood}
  \begin{split}
                    \mbx & \sim p_{\mbtheta}(\mbx \g \mbz) \\
    \Leftrightarrow \mbx & = \cF_{\mbtheta}(\mbxi \semi \mbz), \quad 
                    \mbxi \sim p(\mbxi).    
  \end{split}
\end{equation}
We shall restrict our attention to \emph{prescribed} likelihoods, 
where evaluation of their density is tractable. 
This requires that $\cF_{\mbtheta}(\cd \semi \mbz)$ be a diffeomorphism 
w.r.t. $\mbxi$ and density $p(\mbxi)$ be tractable. 
For example, when $\cF_{\mbtheta}(\cd \semi \mbz)$ is a location-scale 
transform of noise $\mbxi$ and $p(\mbxi)$ is Gaussian, we recover 
a class of familiar Gaussian observation models.
In \cref{sec:recovering_common_latent_variable_models}, we give 
concrete examples of common \acrlongpl{LVM} that can be instantiated 
under this definition.

\subsection{Implicit Prior}
\label{sub:prior}

In \glspl{LVM}, the prior is almost invariably specified as a 
prescribed distribution, most typically a factorized Gaussian centered 
at zero. 
Oftentimes, however, the practitioner possesses prior knowledge that 
simply cannot be embodied within a prescribed distribution. 
To address this limitation, we introduce \emph{implicit} \glspl{LVM}, 
wherein the prior over latent variables is specified as an implicit 
distribution $p^*(\mbz)$, given only by a finite collection $\mbZ^* = 
\{ \mbz_m^* \}_{m=1}^M$ of its samples,
\begin{equation}
  \mbz_m^* \sim p^*(\mbz).
\end{equation}
This formulation offers the utmost degree of flexibility in 
the treatment of prior information, the difficulties of which have
hindered the application of Bayesian statistics since the time of 
 Laplace \citep{Jaynes1968}.

\subsection{Example: Unpaired Image-to-Image Translation}
\label{sub:example_unpaired_image_to_image_translation}

Suppose we have collections of images $\mbX$ and $\mbZ^*$, which are 
assumed to be draws from the data distribution $q^*(\mbx)$ and 
implicit prior distribution $p^*(\mbz)$, respectively. For example,
these might be photographs of natural landscapes and the paintings of 
Van Gogh.

The goal of unpaired image-to-image translation is to learn the
correspondence between variables $\mbx$ and $\mbz$ by capturing the
underlying generative process specified by mapping $\cF_{\mbtheta}$. 
This defines the likelihood $p_{\mbtheta}(\mbx\g\mbz)$---a conditional 
density of $\mbx$ given $\mbz$. Continuing with the above example, 
the problem amounts to learning parameters $\mbtheta$ of the mapping
such that this conditional yields photorealistic renderings of scenes 
portrayed in Van Gogh's paintings. 
Furthermore, the resulting posterior on the latent representation
$p_{\mbtheta}(\mbz \g \mbx)$---a conditional density of $\mbz$ given
$\mbx$---should produce renderings of landscape scenery in Van Gogh's 
iconic style.

\todo[inline]{The posterior satisfies this automatically. By Bayes'
              rule, its places its mass on configurations of the 
              latent variables that explain, or generate, the observed
              data (photorealistic images), and configurations that 
              match the support of the implicit prior (samples of Van 
              Gogh paintings)

              The parameters are learned maximum likelihood estimation,
              by marginalizing out the latent variables and maximizing
              the marginal likelihood}

Parameter learning and inference in \glspl{ILVM} is paved with 
intractabilities. 
For moderately complicated likelihoods, the marginal likelihood 
$p_{\mbtheta}(\mbx)$ is intractable, making it infeasible to perform 
maximum likelihood estimation of $\mbtheta$ and to compute the 
posterior exactly.
Furthermore, the intractability of the prior renders most existing 
approximate inference methods futile. The remainder of this paper 
will be devoted to accurate and scalable approximate inference in 
\glspl{ILVM}.

\section{Variational Inference with Implicit Priors}
\label{sec:variational_inference_with_implicit_priors}

In this section, we describe the first component of our bipartite 
\gls{VI} framework. 
In traditional \gls{VI}, one specifies a family $\cQ$ of densities 
over the latent variables and seeks the member $q \in \cQ$ closest in 
\gls{KL} divergence to the exact posterior $p_{\mbtheta}(\mbz\g\mbx)$
\citep{Jordan1999,Wainwright2008,doi:10.1080/01621459.2017.1285773}. 

\subsection{Prescribed Variational Posterior}
\label{sub:prescribed_variational_posterior}

We begin by describing the variational family $q \in \cQ$. We adopt 
the common practice of \emph{amortizing} inference using an inference 
network \cite{gershman2014amortized}. 
Namely, instead of approximating the exact posterior 
$p_{\mbtheta}(\mbz\g\mbx_n)$ for each $\mbx_n$, using a separate 
variational distribution $q(\mbz; \mblambda_n)$ with local variational 
parameters $\mblambda_n$, we condition on $\mbx$ and optimize a single 
set of variational parameters $\mbphi$ across all $\mbx \sim q^*(\mbx)$.

The variational distribution is then denoted as 
$q_{\mbphi}(\mbz \g \mbx)$, and specified through an inverse mapping 
$\cG_{\mbphi}$ that takes as input random noise 
$\mbepsilon$ and observed variable $\mbx$,
\begin{equation}
  \begin{split}
                    \mbz &\sim q_{\mbphi}(\mbz \g \mbx) \\
    \Leftrightarrow \mbz &= \cG_{\mbphi}(\mbepsilon \semi \mbx), \quad 
                    \mbepsilon \sim p(\mbepsilon).  
  \end{split}
\end{equation}
Just as mapping $\cF_{\mbtheta}$ underpins the generative model, 
mapping $\cG_{\mbphi}$ underpins the \emph{recognition model} 
\citep{DBLP:journals/neco/DayanHNZ95}.
As with the likelihood, we restrict our attention to prescribed 
variational distributions.

As depicted in \cref{fig:graphical_representation_amortized}, the 
dependency relationship between the variational parameters and the 
latent variables now mirrors that of the model parameters and observed 
variables. This symmetry is crucial to the derivation of 
\gls{CYCLEGAN} later in 
\cref{ssub:recovering_kl_through_alternative_divergence_minimization_losses}.

\subsection{Reverse KL Variational Objective}
\label{sub:reverse_kl_variational_objective}

Minimizing the reverse \gls{KL} between the exact and variational 
posterior is equivalent to maximizing the \emph{\gls{ELBO}}, or 
minimizing its \emph{negative}, defined as
\begin{equation}
  \label{eq:nelbo}
  \begin{split}
    \cL_{\textsc{nelbo}}(\mbtheta,\mbphi)
    & \defeq
    \E_{q^*(\mbx) q_{\mbphi}(\mbz\g\mbx)} 
    [ - \log p_{\mbtheta}(\mbx\g\mbz) ] \\
    & \qquad + \E_{q^*(\mbx)} \KL{q_{\mbphi}(\mbz\g\mbx)}{p^*(\mbz)}.    
  \end{split}
\end{equation}
The first term of the \gls{ELBO} is the (negative) \gls{ELL}, 
defined as
\begin{equation}
  \label{eq:nell}
  \cL_{\textsc{nell}}(\mbtheta,\mbphi) 
  \defeq
  \E_{q^*(\mbx) q_{\mbphi}(\mbz\g\mbx)} 
  [ - \log p_{\mbtheta}(\mbx\g\mbz) ].
\end{equation}
It is easy to perform stochastic gradient-based optimization of this 
term by applying the reparameterization trick 
\cite{pmlr-v32-rezende14,Kingma2013}, 
\begin{equation}
  \label{eq:reparameterization}
  \cL_{\textsc{nell}}(\mbtheta,\mbphi) 
  = 
  \E_{q^*(\mbx) p(\mbepsilon)} 
  [ - \log p_{\mbtheta}(\mbx \g \cG_{\mbphi}(\mbepsilon \semi \mbx)) ].
\end{equation}
However, the second term---the \gls{KL} divergence between 
$q_{\mbphi}(\mbz\g\mbx)$ and implicit prior $p^*(\mbz)$---is not so
straightforward. 
In particular, the \gls{KL} divergence can be expressed as
\begin{equation} \label{eq:kl_post_prior}
  \KL{q_{\mbphi}(\mbz\g\mbx)}{p^*(\mbz)} 
  \defeq
  \E_{q_{\mbphi}(\mbz\g\mbx)} [ \log r^*(\mbz \semi \mbx) ],
\end{equation}
where $r^*(\mbz \semi \mbx)$ is defined as the ratio of densities,
\begin{equation} \label{eq:density_ratio_latent}
  r^*(\mbz \semi \mbx) \defeq q_{\mbphi}(\mbz\g\mbx) / p^*(\mbz).
\end{equation}
The dependence on this density ratio is problematic since the prior 
$p^*(\mbz)$ is implicit and cannot be evaluated directly.
To overcome this, we resort to methods for approximating 
$f$-divergences between implicit distributions, which are inextricably 
tied to \emph{\gls{DRE}} 
\citep{Mohamed2016,sugiyama_suzuki_kanamori_2012}.

\subsection{Approximate Divergence Minimization}
\label{sub:approximate_divergence_minimization}

Although we are primarily interested in estimating the \gls{KL} 
divergence of \cref{eq:kl_post_prior}, we give a generalized 
treatment that is applicable to all $f$-divergences 
\citep{Ali1966,ciszar1967information}.
We denote a generic member of the family of $f$-divergences between 
distributions $p$ and $q$ as $\Df{p}{q} \defeq  \E_{p} [ f ( q / p ) ]$, 
for some convex lower-semicontinuous function $f: \bbR_{+} \to \bbR$.

Leveraging results from convex analysis, \citet{DBLP:journals/tit/NguyenWJ10}
devise a variational lower bound that estimates an $f$-divergence
through samples when either or both of the densities are unavailable. 
\citet{NIPS2016_6066} extend this framework to derive \gls{GAN} 
objectives that minimize arbitrary $f$-divergences.
These results are underpin our methodology, and we restate a variant 
of it here for completeness.
\todo[inline]{Is the codomain of first derivative $f'$ always equal 
              to the domain of convex conjugate $f^{\star}$?}
\begin{theorem}[\citealt{DBLP:journals/tit/NguyenWJ10}]
  \label{thm:f_divergence_variational_lower_bound}
  Let $f^{\star}$ be the convex dual of $f$ and $\cR$ a class of 
  functions with codomains equivalent to the domain of $f'$.
  We have the following lower bound on the $f$-divergence 
  between distributions $p(\mbu)$ and $q(\mbu)$,
  \begin{equation}
    \label{eq:f-divergence-bound}
    \begin{split}
      \Df{p(\mbu)}{q(\mbu)}  
      & \geq \max_{\hat{r} \in \cR} 
      \{
      \E_{q(\mbu)}[ f'(\hat{r}(\mbu)) ] \\ 
      & \qquad - \E_{p(\mbu)} 
      [ f^{\star}(f'(\hat{r}(\mbu))) ] \},    
    \end{split}
  \end{equation}
  where equality is attained when $\hat{r}(\mbu)$ is exactly the true 
  density ratio $\hat{r}(\mbu) = q(\mbu) / p(\mbu)$.
\end{theorem}

Applying \cref{thm:f_divergence_variational_lower_bound} to 
$p^*(\mbz)$ and $q_{\mbphi}(\mbz\g\mbx_n)$ for a given $\mbx_n$, and 
optimizing over a class of functions indexed by parameters 
$\mbomega_n$, we obtain the following lower bound on their divergence,
\begin{equation*}
  \begin{split}
    \Df{p^*(\mbz)}{q_{\mbphi}(\mbz\g\mbx_n)} 
    & \geq
    \max_{\mbomega_n}
    \{ 
      \E_{q_{\mbphi}(\mbz\g\mbx_n)} 
      [ f'(r_{\mbomega_n}(\mbz)) ] \\
      & \qquad - \E_{p^*(\mbz)} 
      [ f^{\star}(f'(r_{\mbomega_n}(\mbz))) ] 
    \}.    
  \end{split}
\end{equation*}
While this provides a way to estimate any $f$-divergence between
implicit prior $p^*(\mbz)$ and variational distribution 
$q_{\mbphi}(\mbz\g\mbx_n)$ with only samples,
it also requires us to optimize a separate density ratio estimator 
with parameters $\mbomega_n$ for each observed $\mbx_n$.
Instead, as with the posterior approximation, we also amortize the 
density ratio estimator by conditioning on $\mbx$ and optimizing a 
single set of parameters $\mbalpha$ across all $\mbx \sim q^*(\mbx)$. 
Accordingly, the estimator becomes $r_{\mbalpha}(\mbz \semi \mbx)$, 
taking also $\mbx$ as input. We now maximize an instance of the 
following generalized objective,
\begin{empheq}[box=\fbox]{equation}
  \begin{split}
    \label{eq:f_divergence_variational_lowe_bound_latent}
    \cL_f^{\textnormal{latent}}(\mbalpha \g \mbphi)
    & \defeq 
    \E_{q^*(\mbx) q_{\mbphi}(\mbz\g\mbx)} 
    [ f'(r_{\mbalpha}(\mbz \semi \mbx)) ] \\
    & \quad
    - \E_{q^*(\mbx) p^*(\mbz)} 
    [ f^{\star}(f'(r_{\mbalpha}(\mbz \semi \mbx))) ].    
  \end{split}
\end{empheq} 
\begin{corollary} \label{col:dre_latent_general}
  We have the lower bound,
  \begin{equation} \label{eq:dre_latent_general}
    \E_{q^*(\mbx)} \Df{p^*(\mbz)}{q_{\mbphi}(\mbz\g\mbx)}
    \geq \max_{\mbalpha} \cL_f^{\textnormal{latent}}(\mbalpha \g \mbphi),
  \end{equation}
  with equality at $r_{\mbalpha}(\mbz \semi \mbx) = r^*(\mbz \semi \mbx)$.
\end{corollary}  
\todo[inline]{This is not obvious and requires proof.}
\parhead{Density ratio estimation objective.}
We write $\cL_f^{\textnormal{latent}}(\mbalpha \g \mbphi)$ to denote 
the \gls{DRE} objective, wherein $\mbphi$ is fixed, while $\mbalpha$ 
is a free parameter that varies as this objective is \emph{maximized},
thus tightening the bound of \cref{eq:dre_latent_general} and the 
estimate of the density ratio $r_{\mbalpha}(\mbz \semi \mbx)$. 

\parhead{Divergence minimization loss.}
Conversely, the \emph{\gls{DM}} loss, denoted as 
$\cL_f^{\textnormal{latent}}(\mbphi \g \mbalpha)$, is \emph{minimized} 
w.r.t. $\mbphi$ while $\mbalpha$ remains fixed, thus approximately 
minimizing the $f$-divergence.
In theory, this should be symmetric to the \gls{DRE} objective,
$\cL_f^{\textnormal{latent}}(\mbphi \g \mbalpha) \defeq 
\cL_f^{\textnormal{latent}}(\mbalpha \g \mbphi)$.
However, alternative settings are often used in practice to alleviate 
the problem of vanishing gradients, as we shall see in 
\cref{sec:deriving_cyclegan_as_a_special_case}. 

By applying \cref{col:dre_latent_general} for the setting 
$f_{\textsc{kl}}(u) \defeq u \log u$, we instantiate a lower bound on 
the \gls{KL} divergence of \cref{eq:kl_post_prior} in the following 
objective,
\begin{equation} \label{eq:kl_variational_lower_bound}
  \begin{split}
    \cL_{\textsc{kl}}^{\textnormal{latent}}(\mbalpha \g \mbphi)
    & \defeq 
    \E_{q^*(\mbx) q_{\mbphi}(\mbz\g\mbx)} 
    [ \log r_{\mbalpha}(\mbz \semi \mbx) ] \\
    & \qquad
    - \E_{q^*(\mbx) p^*(\mbz)} [ 
      r_{\mbalpha}(\mbz \semi \mbx) - 1
    ].
  \end{split}
\end{equation}
As we discuss in \cref{sec:relation_to_kliep}, maximization of the
objective in \cref{eq:kl_variational_lower_bound} is closely related 
to the \emph{\gls{KLIEP}} \citep{Sugiyama2008}.

Now, we define the \gls{DM} loss symmetrically to the \gls{DRE} 
objective in \cref{eq:kl_variational_lower_bound}---terms not 
involving $\mbphi$ are omitted,
\begin{align}
  \label{eq:kl_minimization_loss}
  \cL_{\textsc{kl}}^{\textnormal{latent}}(\mbphi \g \mbalpha)
  & \defeq 
  \E_{q^*(\mbx) q_{\mbphi}(\mbz\g\mbx)} [ 
    \log r_{\mbalpha}(\mbz \semi \mbx) 
  ] \\
  & \simeq
  \E_{q^*(\mbx)} \KL{q_{\mbphi}(\mbz\g\mbx)}{p^*(\mbz)}. \nonumber
\end{align}
Combined with the \gls{ELL}, this estimate of the \gls{KL} divergence 
yields an approximation to the \gls{ELBO} where all terms are 
tractable. 
These objectives are summarized in the bi-level optimization problem 
below,
\begin{subequations}
  \label{eq:nelbo_bilevel}
  \begin{align}
    \max_{\mbalpha} \qquad & 
    \cL_{\textsc{kl}}^{\textnormal{latent}}(\mbalpha \g \mbphi), 
    \label{eq:nelbo_bilevel_max} \\
    \min_{\mbphi,\mbtheta} \qquad &
    \cL_{\textsc{kl}}^{\textnormal{latent}}(\mbphi \g \mbalpha)
    +
    \cL_{\textsc{nell}}(\mbtheta,\mbphi),
    \label{eq:nelbo_bilevel_min}
  \end{align}
\end{subequations}
thus concluding the reverse \gls{KL} minimization component of our 
\gls{VI} framework.

\section{Symmetric Joint-Matching Variational Inference}
\label{sec:symmetric_joint_matching_variational_inference}

We now complete the remaining component of our \gls{VI} framework.
In the previous section, we gave an extension to classical \gls{VI},
which is fundamentally concerned with approximating the exact 
posterior. Now, let us instead consider directly approximating the 
\emph{exact joint} $p_{\mbtheta}(\mbx,\mbz)$ through a 
\emph{variational joint} $q_{\mbphi}(\mbx,\mbz)$.

\subsection{Variational Joint}
\label{sub:variational_joint}

Recall that $q^*(\mbx)$ denotes the empirical data distribution.
We define a variational approximation to the exact joint distribution 
of \cref{eq:joint} as
\begin{equation}
  q_{\mbphi}(\mbx,\mbz)
  \defeq
  q_{\mbphi}(\mbz\g\mbx) q^*(\mbx).
\end{equation}
We approximate the exact joint by seeking a variational joint closest 
in \emph{symmetric} \gls{KL} divergence, 
$\KLSYMM{p_{\mbtheta}(\mbx,\mbz)}{q_{\mbphi}(\mbx,\mbz)}$, where
\begin{equation} \label{eq:symm_kl}
  \KLSYMM{p}{q} \defeq \KL{p}{q} + \KL{q}{p}.
\end{equation}

We first look at the reverse \gls{KL} divergence ($\KL{q}{p}$) term. 
When expanded, we see that it is equivalent to the negative \gls{ELBO}
up to additive constants,
\begin{align} 
  \begin{split}
    \label{eq:kl_joint_reverse}
    \textsc{kl}
    & \left [
        q_{\mbphi} (\mbx,\mbz) 
        \,\|\, 
        p_{\mbtheta}(\mbx,\mbz) 
      \right ] \\
    & \defeq \E_{q_{\mbphi}(\mbx,\mbz)} 
        \left [ 
          \log q_{\mbphi}(\mbx,\mbz) 
          - 
          \log p_{\mbtheta}(\mbx,\mbz)
        \right ] 
  \end{split} \\
  & = \cL_{\textsc{nelbo}}(\mbtheta,\mbphi) - \bbH[q^*(\mbx)],
\end{align}
where $\bbH[q^*(\mbx)] \defeq \E_{q^*(\mbx)} [- \log q^*(\mbx)]$ is 
the entropy of $q^*(\mbx)$, a constant w.r.t. parameters $\mbtheta$ 
and $\mbphi$. Hence, minimizing the \gls{KL} divergence of 
\cref{eq:kl_joint_reverse} can be reduced to minimizing 
$\cL_{\textsc{nelbo}}(\mbtheta,\mbphi)$ of \cref{eq:nelbo}, 
without modification.

\subsection{Forward KL Variational Objective}

As for the forward \gls{KL} divergence ($\KL{p}{q}$) term, we have a
similar expansion,
\begin{align}
  \begin{split}
    \label{eq:kl_joint_forward}
    \textsc{kl}
    & \left [
        p_{\mbtheta}(\mbx,\mbz)
        \,\|\, 
        q_{\mbphi} (\mbx,\mbz)
      \right ] \\
    & \defeq \E_{p_{\mbtheta}(\mbx,\mbz)} 
        \left [ 
          \log p_{\mbtheta}(\mbx,\mbz)
          - 
          \log q_{\mbphi}(\mbx,\mbz) 
        \right ]
  \end{split} \\
  \begin{split}
    \label{eq:kl_joint_forward_expansion}
    & = \E_{p^*(\mbz) p_{\mbtheta}(\mbx\g\mbz)}
        [ \log p_{\mbtheta}(\mbx\g\mbz) -
          \log q_{\mbphi}(\mbx,\mbz) ] \\
    & \qquad - \bbH[p^*(\mbz)].
  \end{split}
\end{align}
In analogy with the \gls{ELBO}, we introduce a new variational 
objective that is minimized when the forward KL divergence of
\cref{eq:kl_joint_forward} is minimized. First we define the 
recognition model analog to the marginal likelihood---the 
\emph{marginal posterior}, or \emph{aggregate posterior}, given by
$q_{\phi}(\mbz) \defeq \int q_{\phi}(\mbz\g\mbx) q^*(\mbx) d\mbx$.
It can be approximated by the \emph{\gls{APLBO}}. For consistency, 
we give its \emph{negative}, written as
\begin{equation}
  \label{eq:npalbo}
  \begin{split}
    \cL_{\textsc{naplbo}}(\mbtheta,\mbphi)
    & \defeq
    \E_{p^*(\mbz) p_{\mbtheta}(\mbx\g\mbz)} 
    [ - \log q_{\mbphi}(\mbz\g\mbx) ] \\
    & \quad + \E_{p^*(\mbz)} \KL{p_{\mbtheta}(\mbx\g\mbz)}{q^*(\mbx)}.
  \end{split}
\end{equation}
Furthermore, minimizing the \gls{KL} divergence of 
\cref{eq:kl_joint_forward} can be reduced to minimizing 
$\cL_{\textsc{naplbo}}(\mbtheta,\mbphi)$,
\begin{equation*}
  \textsc{kl}
   \left [
      p_{\mbtheta}(\mbx,\mbz)
      \,\|\, 
      q_{\mbphi} (\mbx,\mbz)
    \right ] 
   = \cL_{\textsc{naplbo}}(\mbtheta,\mbphi) - \bbH[p^*(\mbz)].
\end{equation*}

The first term of the negative \gls{APLBO} is the (negative) 
\emph{\gls{ELP}}, defined as
\begin{equation}
  \label{eq:nelp}
  \cL_{\textsc{nelp}}(\mbtheta,\mbphi) 
  \defeq
  \E_{p^*(\mbz) p_{\mbtheta}(\mbx\g\mbz)} 
  [ - \log q_{\mbphi}(\mbz\g\mbx) ].
\end{equation}
We emphasize a key advantage of having considered the \gls{KL} between 
the joint distributions instead of between the \emph{posteriors}.
Computing the \emph{forward} \gls{KL} divergence between the exact and 
approximate \emph{posterior} distribution is problematic, since it 
requires evaluating expectations over the exact posterior 
$p_{\mbtheta}(\mbz\g\mbx)$, the intractability of which is the reason
we appealed to approximate inference in the first place.

In contrast, the forward \gls{KL} divergence between the exact and
approximate \emph{joint} poses no such difficulties---we are able to 
sidestep the dependency on the exact posterior by expanding it into 
the form of \cref{eq:kl_joint_forward_expansion}. 
Furthermore, as with the \gls{ELBO}, we can perform stochastic 
gradient-based optimization of the \gls{ELP} term by applying the 
same reparameterization trick as in \cref{eq:reparameterization}.

Now, the \gls{KL} divergence term of the \gls{APLBO} in 
\cref{eq:npalbo} can also be expressed as the expected logarithm of a 
density ratio
$r^*(\mbx \semi \mbz) \defeq p_{\mbtheta}(\mbx\g\mbz) / q^*(\mbx)$
that involves an intractable density $q^*(\mbx)$---the empirical data 
distribution.
To overcome this, we adopt the same approach as outlined in
\cref{sub:approximate_divergence_minimization}. 
Namely, we apply \cref{thm:f_divergence_variational_lower_bound} to 
$q^*(\mbx)$ and $p_{\mbtheta}(\mbx\g\mbz^*)$, and fit an amortized
density ratio estimator $r_{\mbbeta}(\mbx \semi \mbz)$ to 
$r^*(\mbx \semi \mbz)$ by maximizing an instance of the generalized 
objective,
\begin{empheq}[box=\fbox]{equation}
  \label{eq:f_divergence_variational_lowe_bound_observed}
  \begin{split}
    \cL_f^{\textnormal{observed}}(\mbbeta \g \mbtheta)
    & \defeq 
    \E_{p^*(\mbz) p_{\mbtheta}(\mbx\g\mbz)} 
    [ f'(r_{\mbbeta}(\mbx \semi \mbz)) ] \\
    & \;
    - \E_{p^*(\mbz) q^*(\mbx)} 
    [ f^{\star}(f'(r_{\mbbeta}(\mbx \semi \mbz))) ].
  \end{split}
\end{empheq} 
\begin{corollary}
  \label{col:dre_observed_general}
  We have the lower bound, 
  \begin{equation}
    \label{eq:dre_observed_general}
    \E_{p^*(\mbz)} \Df{q^*(\mbx)}{p_{\mbtheta}(\mbx\g\mbz)}
    \geq \max_{\mbbeta} \cL_f^{\textnormal{observed}}(\mbbeta \g \mbtheta),
  \end{equation}
  with equality at $r_{\mbbeta}(\mbx \semi \mbz) = r^*(\mbx \semi \mbz)$.
\end{corollary}  
By applying \cref{col:dre_observed_general} with the previously 
defined $f_{\textsc{kl}}(u)$, we obtain lower bound objective 
$\cL_{\textsc{kl}}^{\textnormal{observed}}(\mbbeta \g \mbtheta)$
on the \gls{KL} divergence term in \cref{eq:npalbo}, and a 
corresponding \gls{DM} loss 
$\cL_{\textsc{kl}}^{\textnormal{observed}}(\mbtheta \g \mbbeta)$,
analogous to the definitions of
$\cL_{\textsc{kl}}^{\textnormal{latent}}(\mbalpha \g \mbphi)$ and
$\cL_{\textsc{kl}}^{\textnormal{latent}}(\mbphi \g \mbalpha)$ 
in \cref{eq:kl_variational_lower_bound,eq:kl_minimization_loss},
respectively. See \cref{tab:f_variational_lower_bounds} for a summary
of explicit definitions.

Hence, in addition to the bi-level optimization problems of  
\cref{eq:nelbo_bilevel} we have,
\begin{subequations} 
  \label{eq:naplbo_bilevel}
  \begin{align}
    \max_{\mbbeta} \qquad & 
    \cL_{\textsc{kl}}^{\textnormal{observed}}(\mbbeta \g \mbtheta), 
    \label{eq:naplbo_bilevel_max} \\
    \min_{\mbphi,\mbtheta} \qquad &
    \cL_{\textsc{kl}}^{\textnormal{observed}}(\mbtheta \g \mbbeta)
    +
    \cL_{\textsc{nelp}}(\mbtheta,\mbphi). \label{eq:naplbo_bilevel_min}
  \end{align}
\end{subequations}
As shown, the minimizations in \cref{eq:nelbo_bilevel_min,eq:naplbo_bilevel_min} 
corresponds to minimization of the symmetric \gls{KL} over the joints
$\KLSYMM{p_{\mbtheta}(\mbx,\mbz)}{q_{\mbphi}(\mbx,\mbz)}$, while the
maximizations in \cref{eq:nelbo_bilevel_max,eq:naplbo_bilevel_max} 
approximates the divergences, or more precisely, the density ratios
involving implicit distributions.

\section{CycleGAN as a Special Case}
\label{sec:deriving_cyclegan_as_a_special_case}

In this section, we demonstrate that \acrfull{CYCLEGAN} methods 
\citep{pmlr-v70-kim17a,Zhu2017} can be instantiated 
under our proposed \gls{VI} framework.

\subsection{Basic CycleGAN Framework}
\label{sub:basic_cyclegan_framework}

To address the problem of unpaired image-to-image translation as 
described in \cref{sub:example_unpaired_image_to_image_translation}, 
the \acrshort{CYCLEGAN} model learns two mappings 
$\mbmu_{\mbtheta}: \mbz \mapsto \mbx$ and 
$\mbm_{\mbphi}: \mbx \mapsto \mbz$ by optimizing two complementary 
types of objectives.

\parhead{Distribution matching.}
The first are the adversarial objectives, which help match the output
of mapping $\mbmu_{\mbtheta}$ to the empirical distribution
$q^*(\mbx)$, and the output of $\mbm_{\mbphi}$ to $p^*(\mbz)$.
In particular, for mapping $\mbm_{\mbphi}$, this involves introducing 
a discriminator $\mbD_{\mbalpha}$ and maximizing an adversarial 
objective w.r.t. parameters $\mbalpha$,
\begin{equation}
  \begin{split}
    \label{eq:cyclegan_reverse_gan}
    \ell_{\textsc{gan}}^{\textnormal{reverse}}(\mbalpha \g \mbphi)
    & \defeq 
    \E_{p^*(\mbz)} 
    [ \log \mbD_{\mbalpha}(\mbz) ] \\
    & \quad + \E_{q^*(\mbx)} 
    [ \log(1 - \mbD_{\mbalpha}(\mbm_{\mbphi}(\mbx))) ],
  \end{split}
\end{equation}
while minimizing it w.r.t. parameters $\mbphi$. This encourages 
$\mbm_{\mbphi}$ to produce realistic outputs 
$\mbz = \mbm_{\mbphi}(\mbx), \mbx \sim q^*(\mbx)$ which, to the 
discriminator $\mbD_{\mbalpha}$, are ``indistinguishable'' from 
$\mbz^* \sim p^*(\mbz)$.
An adversarial objective 
$\ell_{\textsc{gan}}^{\textnormal{forward}}(\mbbeta \g \mbtheta)$ is
defined for mapping $\mbmu_{\mbtheta}$ in like manner,
\begin{equation}
  \begin{split}
    \label{eq:cyclegan_forward_gan}
    \ell_{\textsc{gan}}^{\textnormal{forward}}(\mbbeta \g \mbtheta)
    & \defeq 
    \E_{p^*(\mbx)} 
    [ \log \mbD_{\mbbeta}(\mbx) ] \\
    & \quad + \E_{p^*(\mbz)} 
    [ \log(1 - \mbD_{\mbbeta}(\mbmu_{\mbtheta}(\mbz))) ].
  \end{split}
\end{equation}

\parhead{Cycle-consistency.} The second type are the cycle-consistency 
losses, which enforce tight correspondence between domains by ensuring 
that reconstruction $\mbx' = \mbmu_{\mbtheta}(\mbm_{\mbphi}(\mbx))$ 
is close to the input $\mbx$, 
and likewise for $\mbm_{\mbphi}(\mbmu_{\mbtheta}(\mbz))$. This is 
achieved by minimizing a reconstruction loss,
\begin{equation}
  \label{eq:cyclegan_reverse_consistency}
  \ell_{\textsc{const}}^{\textnormal{reverse}}(\mbtheta,\mbphi)
  \defeq 
  \E_{q^*(\mbx)} 
  [ \| \mbx - \mbmu_{\mbtheta}(\mbm_{\mbphi}(\mbx)) \|_{\rho}^{\rho} ],
\end{equation}
where $\| \cdot \|_{\rho}$ denotes the $\ell_{\rho}$-norm. 
A similar loss $\ell_{\textsc{const}}^{\textnormal{forward}}(\mbtheta,\mbphi)$ 
is defined for the reconstruction of $\mbz$,
\begin{equation}
  \label{eq:cyclegan_forward_consistency}
  \ell_{\textsc{const}}^{\textnormal{forward}}(\mbtheta,\mbphi)
  \defeq 
  \E_{p^*(\mbz)} 
  [ \| \mbz - \mbm_{\mbphi}(\mbmu_{\mbtheta}(\mbz)) \|_{\rho}^{\rho} ].
\end{equation}
These objectives are summarized in the following set of optimization 
problems,
\begin{subequations}
\label{eq:cyclegan_bilevel}
\begin{align}
  \max_{\mbalpha} \; & 
  \ell_{\textsc{gan}}^{\textnormal{reverse}}(\mbalpha \g \mbphi), 
  \quad 
  \max_{\mbbeta} \; 
  \ell_{\textsc{gan}}^{\textnormal{forward}}(\mbbeta \g \mbtheta), \label{eq:cyclegan_bilevel_max} \\
  \min_{\mbphi,\mbtheta} \; &
  \ell_{\textsc{gan}}^{\textnormal{reverse}}(\mbphi \g \mbalpha)
  + \ell_{\textsc{const}}^{\textnormal{reverse}}(\mbtheta,\mbphi), \label{eq:cyclegan_bilevel_min_reverse} \\
  \min_{\mbphi,\mbtheta} \; &
  \ell_{\textsc{gan}}^{\textnormal{forward}}(\mbtheta \g \mbbeta) 
  + \ell_{\textsc{const}}^{\textnormal{forward}}(\mbtheta,\mbphi). \label{eq:cyclegan_bilevel_min_forward} 
\end{align}  
\end{subequations}
We now highlight the correspondences between these objectives and 
those of our proposed \gls{VI} framework, as summarized in the 
optimization problems of \cref{eq:naplbo_bilevel,eq:nelbo_bilevel}.

\subsection{Cycle-consistency as Conditional Probability Maximization}
\label{sub:cycle_consistency_as_conditional_probability_maximization}

We now demonstrate that minimizing the cycle-consistency losses 
corresponds to maximizing the expected log likelihood and 
variational posterior of \cref{eq:nell,eq:nelp}.
This can be shown by instantiating specific classes of 
$p_{\mbtheta}(\mbx \g \mbz)$ and $q_{\mbphi}(\mbz \g \mbx)$ that recover
$\ell_{\textsc{const}}^{\textnormal{reverse}}(\mbtheta,\mbphi)$ and
$\ell_{\textsc{const}}^{\textnormal{forward}}(\mbtheta,\mbphi)$ from
$\cL_{\textsc{nell}}(\mbtheta,\mbphi)$ and 
$\cL_{\textsc{nelp}}(\mbtheta,\mbphi)$, respectively.

\begin{proposition}
  \label{prop:consistency_expected_conditional}
  Consider a typical case where the likelihood and the  posterior 
  approximation are both Gaussians,
  \begin{align*}
    p_{\mbtheta}(\mbx \g \mbz) 
    & \defeq 
    \cN(\mbx \g \mbmu_{\mbtheta}(\mbz), \tau^2 \mbI), \\
    q_{\mbphi}(\mbz \g \mbx) 
    & \defeq 
    \cN(\mbz \g \mbm_{\mbphi}(\mbx), t^2 \mbI).
  \end{align*}
  Then, in the limit as the posterior variance tends to zero, 
  $\cL_{\textsc{nell}}(\mbtheta,\mbphi)$
  approaches
  $\ell_{\textsc{const}}^{\textnormal{reverse}}(\mbtheta,\mbphi)$
  for $\rho=2$, up to constants\footnote{we obtain the same result for 
  the case $\rho = 1$ by instead setting both the likelihood and 
  approximate posterior to be Laplace distributions.}. 
  Formally stated,
  \begin{align*}
    \cL_{\textsc{nell}}(\mbtheta,\mbphi) 
    & \to
    \gamma_1 \ell_{\textsc{const}}^{\textnormal{reverse}}(\mbtheta,\mbphi) + \delta_1
    \quad \textnormal{as } t \to 0 \\
    & \propto \ell_{\textsc{const}}^{\textnormal{reverse}}(\mbtheta,\mbphi). 
  \end{align*} 
  where $\gamma_1 = \frac{1}{2\tau^2}$ and 
  $\delta_1 = \frac{D}{2} \log \frac{\pi}{\gamma_1}$.
  Likewise,
  \begin{align*}
    \cL_{\textsc{nelp}}(\mbtheta,\mbphi) 
    & \to
    \gamma_2 \ell_{\textsc{const}}^{\textnormal{forward}}(\mbtheta,\mbphi) + \delta_2
    \quad \textnormal{as } \tau \to 0 \\
    & \propto \ell_{\textsc{const}}^{\textnormal{forward}}(\mbtheta,\mbphi). 
  \end{align*} 
  where $\gamma_2 = \frac{1}{2t^2}$ and 
  $\delta_2 = \frac{K}{2} \log \frac{\pi}{\gamma_2}$.
\end{proposition}

The proof is given in 
\cref{sec:proof_consistency_expected_conditional}.
Hence, roughly speaking, the cycle-consistency losses can be seen as
specific cases of the \gls{ELL} and \gls{ELP} with \emph{degenerate} 
conditional distributions.
Furthermore, this sheds new light on the roles of the cycle-consistency 
losses. In particular, the reverse consistency loss---like the \gls{ELL} 
term---encourages the conditional $q_{\mbphi}(\mbz \g \mbx)$ to place 
its mass on configurations of latent variables that can explain, or in 
this case, \emph{represent} the data well. 

\subsection{Distribution Matching as Approximate Divergence Minimization}
\label{sub:distribution_matching_as_approximate_divergence_minimization}

We now discuss how the adversarial objectives 
$\ell_{\textsc{gan}}^{\textnormal{reverse}}(\mbalpha \g \mbphi)$ and
$\ell_{\textsc{gan}}^{\textnormal{forward}}(\mbbeta \g \mbtheta)$ 
relate to the \gls{KL} variational lower bounds of our framework,
$\cL_{\textsc{kl}}^{\textnormal{latent}}(\mbalpha \g \mbphi)$ and 
$\cL_{\textsc{kl}}^{\textnormal{observed}}(\mbbeta \g \mbtheta)$, 
respectively. To reduce clutter, we restrict our discussion to the 
reverse objective 
$\ell_{\textsc{gan}}^{\textnormal{reverse}}(\mbalpha \g \mbphi)$, as
the same reasoning readily applies to the forward 
$\ell_{\textsc{gan}}^{\textnormal{forward}}(\mbbeta \g \mbtheta)$.

\subsubsection{As Density Ratio Estimation by 
               Probabilistic Classification}

Firstly, the connections between \glspl{GAN}, divergence minimization 
and \gls{DRE} are well-established \citep{Mohamed2016,
NIPS2016_6066,Uehara2016}. Although 
$\ell_{\textsc{gan}}^{\textnormal{reverse}}(\mbalpha \g \mbphi)$ 
is a scoring rule for probabilistic classification 
\citep{10.2307/27639845}, one can readily show that it can also be 
subsumed as an instance of the generalized variational lower bound 
$\cL_f^{\textnormal{latent}}(\mbalpha \g \mbphi)$.
Furthermore, similar to
$\cL_{\textsc{kl}}^{\textnormal{latent}}(\mbalpha \g \mbphi)$,
maximizing
$\ell_{\textsc{gan}}^{\textnormal{reverse}}(\mbalpha \g \mbphi)$ 
corresponds estimating the intractable density ratio $r^*(\mbz \semi \mbx)$ of 
\cref{eq:density_ratio_latent}. 
\begin{lemma}
\label{prop:cyclegan_reverse_gan_general}
By setting
$f_{\textsc{gan}}(u) = u \log u - (u+1) \log (u+1)$ in the generalized 
objective $\cL_f^{\textnormal{latent}}(\mbalpha \g \mbphi)$ of 
\cref{eq:f_divergence_variational_lowe_bound_latent}, we instantiate 
the objective
\begin{equation}
  \label{eq:cyclegan_reverse_gan_general}
  \begin{split}
    \cL_{\textsc{gan}}^{\textnormal{reverse}}(\mbalpha \g \mbphi)
    & \defeq
    \E_{q^*(\mbx) p^*(\mbz)} 
    [ \log \cD_{\mbalpha}(\mbz \semi \mbx) ] \\
    & + \E_{q^*(\mbx) q_{\mbphi}(\mbz\g\mbx)} 
    [ \log (1 - \cD_{\mbalpha}(\mbz \semi \mbx)) ],
  \end{split} 
\end{equation}
where $\cD_{\mbalpha}(\mbz \semi \mbx) \defeq 
1 - \sigma(\log r_{\mbalpha}(\mbz \semi \mbx))$, and $\sigma$ is the 
logistic sigmoid function. 
\end{lemma}
\begin{lemma} 
  \label{prop:cyclegan_reverse_gan_reduction}
  By specifying a discriminator 
  $\cD_{\mbalpha}(\mbz \semi \mbx) = \mbD_{\mbalpha}(\mbz)$
  which ignores auxiliary input $\mbx$, and mapping 
  $\cG_{\mbphi}(\mbepsilon \semi \mbx) = \mbm_{\mbphi}(\mbx)$
  which ignores noise input $\mbepsilon$, 
  $\cL_{\textsc{gan}}^{\textnormal{reverse}}(\mbalpha \g \mbphi)$ 
  reduces to 
  $\ell_{\textsc{gan}}^{\textnormal{reverse}}(\mbalpha \g \mbphi)$.
\end{lemma}
\begin{proposition} 
\label{prop:cyclegan_reverse_gan_as_variational_lower_bound}
The reverse adversarial objective 
$\ell_{\textsc{gan}}^{\textnormal{reverse}}(\mbalpha \g \mbphi)$ 
can be subsumed as an instance of the generalized variational lower bound 
$\cL_f^{\textnormal{latent}}(\mbalpha \g \mbphi)$.
\end{proposition}
\Cref{prop:cyclegan_reverse_gan_as_variational_lower_bound} follows
directly from \cref{prop:cyclegan_reverse_gan_reduction,prop:cyclegan_reverse_gan_general}.
Their proofs are given in \cref{sec:proof_of_cyclegan_reverse_gan_reduction,sec:proof_of_cyclegan_reverse_gan_general}, respectively.
Now, by \cref{col:dre_latent_general}, objective 
$\cL_{\textsc{gan}}^{\textnormal{reverse}}(\mbalpha \g \mbphi)$ 
is maximized exactly when $r_{\mbalpha}(\mbz \semi \mbx) = 
r^*(\mbz \semi \mbx)$. Hence, we can interpret
$\cL_{\textsc{gan}}^{\textnormal{reverse}}(\mbalpha \g \mbphi)$ as an
objective for density ratio estimation based on probabilistic 
classification, while 
$\cL_{\textsc{kl}}^{\textnormal{latent}}(\mbalpha \g \mbphi)$ is an
objective based on \gls{KLIEP}.

Now, the default choice of \gls{DM} loss is
$\cL_{\textsc{gan}_\textsc{a}}^{\textnormal{reverse}}(\mbphi \g \mbalpha)
\defeq \cL_{\textsc{gan}}^{\textnormal{reverse}}(\mbalpha \g \mbphi)$.
Omitting terms not involving $\mbphi$, this is given by
\begin{equation}
  \cL_{\textsc{gan}_\textsc{a}}^{\textnormal{reverse}}(\mbphi \g \mbalpha)
  \defeq \E_{q^*(\mbx) q_{\mbphi}(\mbz\g\mbx)} 
  [ \log(1 - \cD_{\mbalpha}(\mbz \semi \mbx)) ].
\end{equation}
Unlike
$\cL_{\textsc{kl}}^{\textnormal{latent}}(\mbphi \g \mbalpha)$, minimizing
$\cL_{\textsc{gan}_\textsc{a}}^{\textnormal{reverse}}(\mbphi \g \mbalpha)$
does not minimize the \gls{KL} divergence of 
\cref{eq:kl_post_prior}. Hence, the minimization problem of
\cref{eq:cyclegan_bilevel_min_reverse} does not correspond to that of
\cref{eq:nelbo_bilevel_min}, and so does not maximize the \gls{ELBO}, 
or any known \gls{VI} objective.


\subsubsection{Recovering KL Through Alternative Divergence 
               Minimization Losses}
\label{ssub:recovering_kl_through_alternative_divergence_minimization_losses}

Although the default choice of \gls{DM} loss does not yield a tight 
correspondence to \gls{VI}, the existing \gls{CYCLEGAN} frameworks---and 
indeed most \gls{GAN}-based approaches---arbitrarily select an 
alternative \gls{DM} loss that avoids vanishing gradients, and work 
well in practice. Hence, one need only choose an alternative that 
\emph{does} correspond to minimizing the \gls{KL} divergence of 
\cref{eq:kl_post_prior}.

Firstly, of the \gls{CYCLEGAN} methods, \citet{pmlr-v70-kim17a} adopt 
the widely-used \gls{DM} loss originally suggested by \citet{NIPS2014_5423}, 
\begin{equation}
  \cL_{\textsc{gan}_\textsc{b}}^{\textnormal{reverse}}(\mbphi \g \mbalpha)
  \defeq \E_{q^*(\mbx) q_{\mbphi}(\mbz\g\mbx)} 
  [ - \log \cD_{\mbalpha}(\mbz \semi \mbx) ],
\end{equation}
while \citet{Zhu2017} optimize the \gls{LSGAN} objectives of \citet{Mao2016}.

Consider the \emph{combination} of losses 
$\cL_{\textsc{gan}_\textsc{a}}^{\textnormal{reverse}}(\mbphi \g \mbalpha)$
and $\cL_{\textsc{gan}_\textsc{b}}^{\textnormal{reverse}}(\mbphi \g \mbalpha)$,
\begin{align}
  \cL_{\textsc{gan}_\textsc{c}}^{\textnormal{reverse}}(\mbphi \g \mbalpha)
  & \defeq
  \cL_{\textsc{gan}_\textsc{a}}^{\textnormal{reverse}}(\mbphi \g \mbalpha) 
  +
  \cL_{\textsc{gan}_\textsc{b}}^{\textnormal{reverse}}(\mbphi \g \mbalpha) \\
  & = \nonumber
  \E_{q^*(\mbx) q_{\mbphi}(\mbz\g\mbx)} 
  \left [ - \log \frac{\cD_{\mbalpha}(\mbz \semi \mbx)}
                      {1 - \cD_{\mbalpha}(\mbz \semi \mbx)} \right ].
\end{align}
\begin{proposition}  
\label{prop:dm_loss_kl_equivalence}
We have $\cL_{\textsc{gan}_\textsc{c}}^{\textnormal{reverse}}(\mbphi \g \mbalpha) 
= \cL_{\textsc{kl}}^{\textnormal{latent}}(\mbphi \g \mbalpha).$
\end{proposition}
\Cref{prop:dm_loss_kl_equivalence} was originally observed by 
\citet{sonderby2016amortised} and is shown in 
\cref{sec:proof_of_dm_loss_kl_equivalence}.
Thus, for the setting of the \gls{DM} loss
$\cL_{\textsc{gan}_\textsc{c}}^{\textnormal{reverse}}(\mbphi \g \mbalpha)$,
the minimization problem of \cref{eq:cyclegan_bilevel_min_reverse} 
corresponds to that of \cref{eq:nelbo_bilevel_min}, and thus maximizes 
the \gls{ELBO}. 
This is equivalent to fitting the density ratio estimator 
$r_{\mbalpha}(\mbz \semi \mbx)$ by maximizing the objective
$\cL_{\textsc{gan}}^{\textnormal{reverse}}(\mbalpha \g \mbphi)$
instead of 
$\cL_{\textsc{kl}}^{\textnormal{latent}}(\mbalpha \g \mbphi)$, and 
plugging it back into 
$\cL_{\textsc{kl}}^{\textnormal{latent}}(\mbphi \g \mbalpha)$ to 
approximately minimize the \gls{KL} divergence of \cref{eq:kl_post_prior}.
Such an approach is prevalent among existing implicit \gls{VI} methods
\citep{NIPS2017_7136,pmlr-v70-mescheder17a,Huszar2017,NIPS2017_7020}.

In summary, we have that the cycle-consistency losses are a specific 
instance of the \gls{ELL} and \gls{ELP}, while the adversarial 
objectives are a specific instance of the variational lower bound
for divergence estimation, the maximization of which can be seen as 
density ratio estimation by probabilistic classification. 
By explicitly setting the corresponding divergence minimization loss 
such that it leads to minimization of the required \gls{KL} divergence 
terms in the \gls{ELBO} and \gls{APLBO}, we subsume the \gls{CYCLEGAN} 
model under our proposed \gls{VI} framework. See 
\cref{sec:summary_of_definitions} for a succinct summary of the 
relationships.

\section{Related Work}
\label{sec:related_work}

This paper is closely related to the recent works that seek to extend 
the scope of \gls{VI} to implicit distributions, making it feasible in 
scenarios where one or more of the densities that constitute the 
\gls{ELBO} are not explicitly available.
A recurring theme throughout this line of work is approximation of the 
\gls{ELBO} by exploiting the formal connection between density ratio 
estimation and \glspl{GAN} \cite{Uehara2016,Mohamed2016}. The major 
variation is in the choice of the target density ratio being estimated,
which is dictated by the problem setting. 
\citet{pmlr-v70-mescheder17a,Makhzani2015} estimate the density 
ratio $q_{\mbphi}(\mbz\g\mbx) / p(\mbz)$ so as to allow for 
arbitrarily expressive sample-based posterior approximations 
$q_{\mbphi}(\mbz\g\mbx)$. 
This corresponds to the reverse \gls{KL} minimization component of 
our approach, wherein we also estimate the same density ratio, but 
instead to allow for implicit prior distributions $p(\mbz)$.
\newpage
Similar to \acrshort{BIGAN} \cite{Dumoulin2016} and \acrshort{ALI} 
\cite{Donahue2016}, \citet[\acrshort{LFVI}]{NIPS2017_7136} match a 
variational joint to an exact joint distribution by estimating the 
density ratio $p_{\mbtheta}(\mbx,\mbz)/q_{\mbphi}(\mbx,\mbz)$ and 
using it to approximately minimize the \gls{KL} divergence.
Although this formulation relaxes the requirement of having \emph{any} 
tractable densities, their focus is on inference for models with 
intractable likelihoods $p_{\mbtheta}(\mbx \g \mbz)$, and also 
incorporate the implicit posteriors of \acrshort{AVB}. 
In our setting, the joint's intractability is due instead to the 
implicit prior $p^*(\mbz)$. While we also 
approximate the exact joint, we do so by minimizing a \emph{symmetric}
\gls{KL} divergence. Furthermore, since both 
$p_{\mbtheta}(\mbx \g \mbz)$ and 
$q_{\mbphi}(\mbz\g\mbx)$ are prescribed, we incorporate them explicitly 
within our loss functions, and estimate a different set of density ratios.
This closely resembles the approach of \citet{NIPS2017_7020}, which 
also minimizes the symmetric \gls{KL} divergence between the joints. 
However, the focus of their method is not on implicit distributions, 
and thus specify a different set of losses than ours---one that 
requires solving more complicated density ratio estimation problems. 
More importantly, their method does not yield a tight correspondence 
to \gls{CYCLEGAN} models. 

Next, similar to InfoGAN \cite{Chen2016} and \textsc{veegan} 
\cite{srivastava2017veegan}, the forward \gls{KL} minimization 
component of our method also optimizes a model of the latent 
variables, which is reminiscent of the wake-sleep algorithm for 
training Helmholtz machines \cite{DBLP:journals/neco/DayanHNZ95}. 
This is discussed further by \citet{Hu2017}, who provide a 
comprehensive treatment of the links between the work mentioned in 
this section, and importantly, the symmetric perspective of generation 
on recognition that is fundamental to our approach.


\section{Experiments}
\label{sec:experiments}

\begin{figure*}[t]
  \centering
  \begin{subfigure}[t]{.32\textwidth}
    \centering
    \includegraphics[width=\columnwidth]{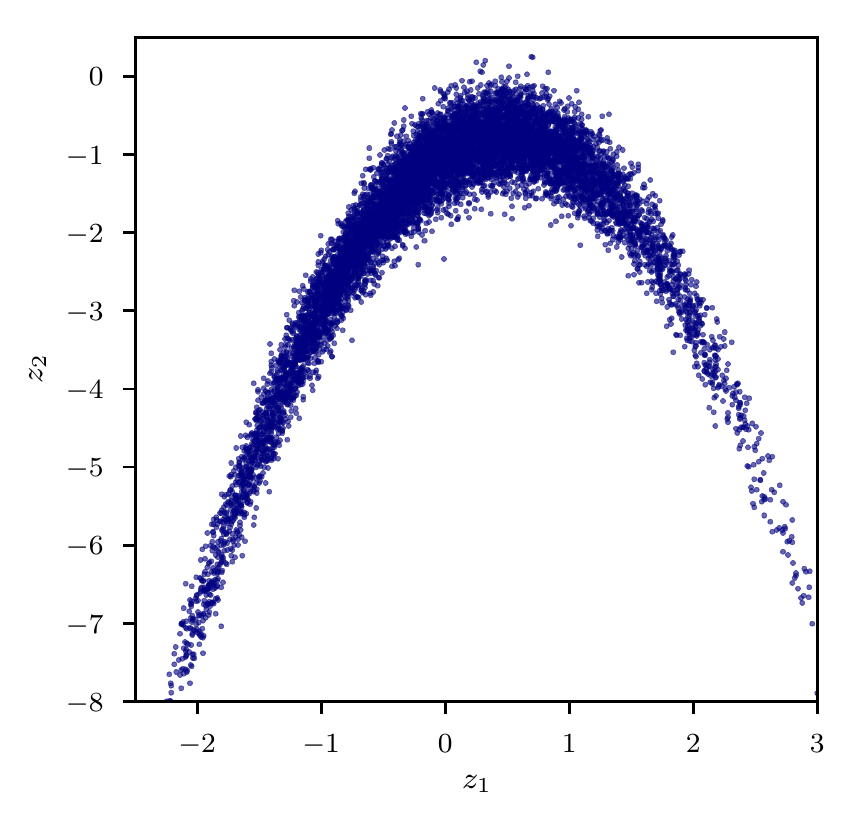}
    \caption{10k samples from implicit prior $p^*(\mbz)$.} \label{fig:prior_samples}
  \end{subfigure}
  \begin{subfigure}[t]{.32\textwidth}
    \centering
    \includegraphics[width=\columnwidth]{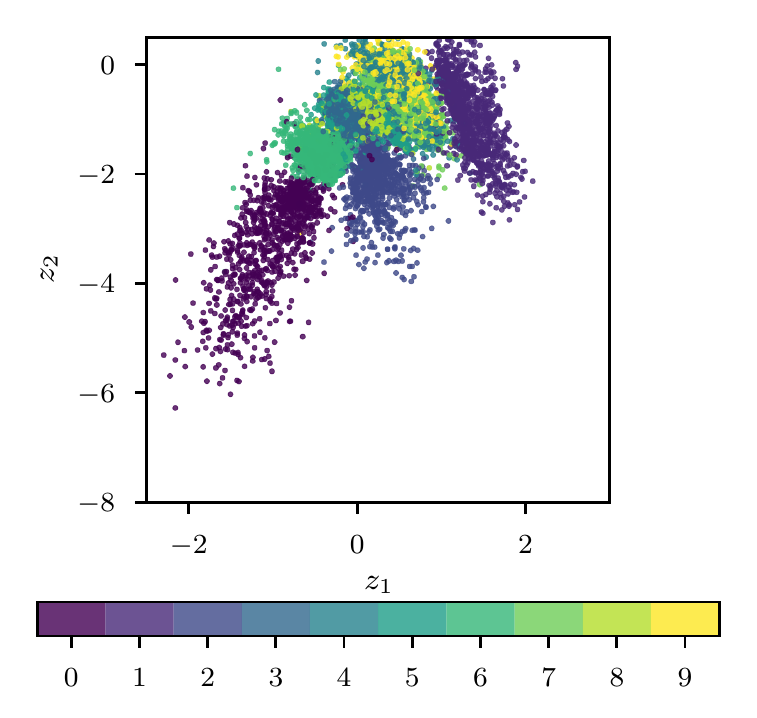}
    \caption{Mean of $q_{\mbphi}(\mbz \g \mbx)$ for every $\mbx$ 
           from held-out test set of size 10k, colored by digit class.}
    \label{fig:posterior_samples}
  \end{subfigure}
  \begin{subfigure}[t]{.32\textwidth}
    \centering
    \includegraphics[width=\columnwidth]{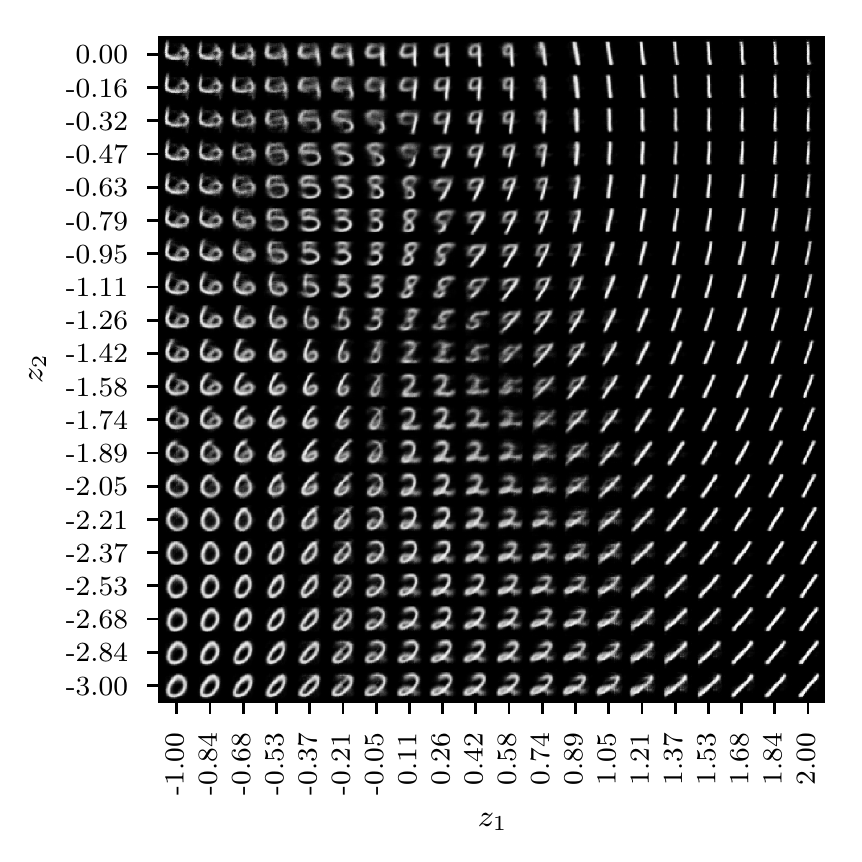}
    \caption{Mean of $p_{\mbtheta}(\mbx \g \mbz)$ for $20 \times 20$
             values of $\mbz$ along a uniform grid.}
    \label{fig:observed_manifold}
  \end{subfigure}
  \caption{Visualization of 2D latent space and the corresponding
           observed space manifold.}
\end{figure*}

To empirically assess the performance of our approach, we consider 
the problem of reducing the dimensionality of the \textsc{mnist} dataset 
to a 2D latent space, wherein the prior distribution on the latent 
representations is specified by its samples (shown in \cref{fig:prior_samples}). 
This ``banana-shaped distribution'' is a commonly used testbed for
adaptive \acrshort{MCMC} methods \citep{Titsias2017,Haario1999}. Its 
samples can be generated by drawing from a bivariate Gaussian 
with unit variances and correlation $\rho=0.95$, and transforming them 
through mapping $H(z_1, z_2) \defeq [z_1, z_2 - z_1^2 - 1]^T$. 
While the density of this distribution can be computed,
it is withheld from our algorithm and used only in the \gls{VAE} 
baseline, which does not permit implicit distributions.
\Cref{fig:posterior_samples} shows, for every observation $\mbx$ from 
a held-out test set, the mean $\mbm_{\mbphi}(\mbx)$ of the posterior 
$q_{\mbphi}(\mbz \g \mbx)$ over its underlying latent representation 
$\mbz$.
\begin{table}[t]
\caption{\Acrlongpl{MSE} of reconstructions.}
\centering
\label{tab:metrics}
\vskip 0.15in
\begin{center}
  \begin{small}
    \begin{sc}
      \begin{tabular}{lcr}
        \toprule
        Method & mse $\mbz$ & mse $\mbx$ \\
        \midrule
        sjmvi (ours) & \textbf{0.17} & \textbf{0.04} \\
        vae \citep{Kingma2013} & 0.88 & \textbf{0.04} \\
        avb \citep{pmlr-v70-mescheder17a} & 0.29 & \textbf{0.04} \\
        \bottomrule
      \end{tabular}
    \end{sc}
  \end{small}
\end{center}
\vskip -0.1in
\end{table}
Observe that instances of the various digit classes are disentangled 
in this latent space, while still closely matching the shape of the 
prior distribution, despite having only access to its samples. The 
resulting manifold of reconstructions is depicted in \cref{fig:observed_manifold}.

In \cref{tab:metrics}, we report the \gls{MSE} on the reconstructions 
of observations from the held-out test set and benchmark against 
\textsc{vae} / \textsc{avb}. 
Also, for the joint approximation to properly match the support of 
the exact joint, the latent codes should also be representable by
its corresponding observation. Hence, we also report the 
\gls{MSE} between samples from the prior and their reconstructions. 
While we find no improvements on reconstruction quality of 
observations, our method significantly outperforms others in 
reconstructing latent codes, suggesting our method has greater 
capacity to faithfully approximate the exact joint.

\section{Conclusion}
\label{sec:conclusion}

We introduced \acrlongpl{ILVM}, which offer the greatest extent of
flexibility in treatment of prior information, and can be used in 
to model problems ranging from dimensionality reduction to unpaired
image-to-image translation.
For their parameter estimation and inference, we developed a 
\acrlong{VI} framework that augments traditional \acrshort{VI} 
approaches by minimizing the symmetric \gls{KL} between the exact 
joint distribution and an approximate distribution.

Additionally, we provided a theoretical treatment of the link 
between \glspl{CYCLEGAN} and approximate Bayesian inference.
In short, samples from the two domains correspond respectively to 
those drawn from the data and implicit prior distribution in a 
\gls{ILVM}. Parameter learning in \glspl{CYCLEGAN} corresponds to
approximate inference in this \gls{ILVM} under our proposed \gls{VI} 
framework. The forward and reverse mappings in \glspl{CYCLEGAN} arise 
naturally in the generative and recognition models,
while the cycle-consistency constraints correspond to their log 
probabilities, and the adversarial losses are approximations to an 
$f$-divergence.

By lifting the requirement of prescribed prior distributions in favor
of arbitrarily flexible implicit distributions, we can discover 
different perspectives on existing learning methods and provide more 
flexible approaches to probabilistic modeling.

\newpage

\section*{Acknowledgements}

We are grateful to Taeksoo Kim, Alistair Reid, Kelvin Hsu, 
Hisham Husain and Harrison Nguyen and anonymous reviewers for 
insightful discussion and feedback.
Louis Tiao is partially supported by the CSIRO Data61 Postgraduate 
Scholarship.

\bibliography{main}
\bibliographystyle{icml2018}

\clearpage

\appendix

\renewcommand\thefigure{\thesection.\arabic{figure}}    

\section{Graphical Representation}
\label{sec:graphical_representation}

\begin{figure}[h]
  \centering
  \begin{subfigure}[t]{0.45\columnwidth}
    \centering
    \begin{tikzpicture}[scale=1, transform shape]
      \node[obs] (x1) {$\mbx_n$};
      \node[latent, above=of x1] (z1) {$\mbz_n$};
      \node[const, left=of z1] (phi1) {$\mblambda_n$};
      \node[const, right=of z1] (theta1) {$\mbtheta$};
      \edge [dashed] {phi1} {z1};
      \edge {theta1} {x1};
      \edge {z1} {x1};
      \plate [xscale=1.1] {} {(x1)(z1)(phi1)} {$N$} ;
    \end{tikzpicture}
    \caption{Without amortized inference, each local latent variable 
             is governed by its own local variational parameters.}
    \label{fig:graphical_representation_not_amortized}
  \end{subfigure}
  \begin{subfigure}[t]{0.45\columnwidth}
    \centering
    \begin{tikzpicture}[scale=1, transform shape]
      \node[obs] (x1) {$\mbx_n$};
      \node[latent, above=of x1] (z1) {$\mbz_n$};
      \node[const, left=of z1] (phi1) {$\mbphi$};
      \node[const, right=of z1] (theta1) {$\mbtheta$};
      \edge [dashed] {phi1} {z1};
      \edge {theta1} {x1};
      \draw (x1) edge[out=135,in=225,->,dashed] (z1);
      \edge {z1} {x1};
      \plate [xscale=1.5] {} {(x1)(z1)} {$N$} ;
    \end{tikzpicture}
    \caption{With amortized inference, we condition on observed 
             variables and employ a single set of global variational 
             parameters.}
    \label{fig:graphical_representation_amortized}
  \end{subfigure}
      \caption{Graphical representation of the \emph{generative model} 
               (\textbf{solid}) and the \emph{recognition model} 
               (\textbf{dashed}).}
    \label{fig:graphical_representation}
\end{figure}
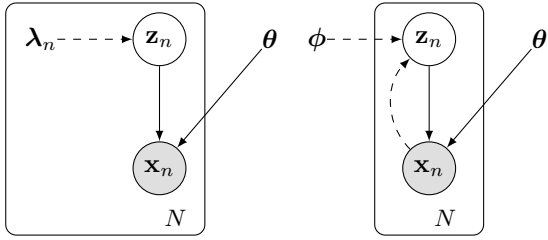

The graphical representation of \acrlongpl{ILVM} is depicted below in 
\cref{fig:graphical_representation}.

\section{Recovering Common Latent Variable Models}
\label{sec:recovering_common_latent_variable_models}

Our model specification is sufficiently general for encapsulating a 
broad range of familiar latent variable models, even when we make 
simplifying assumptions on the mapping $\cF_{\mbtheta}(\cd \semi \mbz)$. 
In particular, consider the special case where the mapping is an 
affine transformation of the noise vector $\mbxi$,
\begin{equation*}
  \cF_{\mbtheta}(\mbxi \semi \mbz) 
  \defeq 
  \mbmu_{\mbtheta}(\mbz) + \mbSigma_{\mbtheta}(\mbz)^{\frac{1}{2}} \mbxi, 
  \quad \mbxi \sim \cN(\mbzero, \mbI),
\end{equation*}
for functions $\mbmu_{\mbtheta}$ and $\mbSigma_{\mbtheta}$ 
parameterized by $\mbtheta$ that take $\mbz$ as input.
To simplify matters further, assume $\mbSigma_{\mbtheta}$ is 
constant w.r.t. to its input, i.e. $\mbSigma_{\mbtheta}(\mbz) = 
\mbPsi$ for all $\mbz$.
The likelihood can then be written explicitly as
\begin{equation*}
  p_{\mbtheta}(\mbx \g \mbz) 
  = 
  \cN(\mbx \g  \mbmu_{\mbtheta}(\mbz), \mbPsi).
\end{equation*}

\parhead{Factor analysis \& probabilistic PCA.}
In the case where the mean function $\mbmu_{\mbtheta}$ is an affine 
transformation of $\mbz$,
\begin{equation*}
  \mbmu_{\mbtheta}(\mbz) \defeq \mbW \mbz + \mbb,
\end{equation*}
and the covariance matrix is diagonal 
$\mbPsi = \diag(\psi_1^2, \dotsc, \psi_D^2)$, we recover 
\emph{\gls{FA}} \citep{bartholomew2011latent}. 
Furthermore, when the covariance matrix is isotropic $\mbPsi = 
\psi^2\mbI$, we recover \emph{\gls{PPCA}} \citep{Tipping1999}. 
In this example, the parameters $\mbtheta$ consist of the  factor 
loading matrix $\mbW$, the bias vector $\mbb$ and the covariance 
matrix $\mbPsi$.  

\parhead{Deep and nonlinear latent variable models.}
By introducing nonlinearities to the mean function, we are able to 
recover nonlinear factor analysis \citep{lappalainen2000bayesian}, 
nonlinear Gaussian sigmoid belief networks \citep{frey1999variational}, 
and other more sophisticated variants of deep latent variable models. 
When the mapping is defined by a \gls{MLP}, we can recover simple 
instances of a \gls{VAE} with a Gaussian probabilistic decoder 
\citep{Kingma2013,pmlr-v32-rezende14}.



\section{Proof of \Cref{prop:consistency_expected_conditional}}
\label{sec:proof_consistency_expected_conditional}

\begin{proof} 
  Firstly, note the generative mappings underlying the given Gaussian 
  likelihood and approximate posterior are
  \begin{equation*}
    \begin{split}
      p_{\mbtheta}(\mbx \g \mbz) 
      & \defeq 
      \cN(\mbx \g \mbmu_{\mbtheta}(\mbz), \tau^2 \mbI), \\
      \Leftrightarrow 
      \cF_{\mbtheta}(\mbxi \semi \mbz) 
      & \defeq \mbmu_{\mbtheta}(\mbz) + \tau \mbxi, \quad 
      \mbxi \sim \cN(\mbzero, \mbI),
    \end{split}
  \end{equation*}
  and,
  \begin{equation*}
    \begin{split}
      q_{\mbphi}(\mbz \g \mbx) 
      & \defeq 
      \cN(\mbz \g \mbm_{\mbphi}(\mbx), t^2 \mbI), \\
      \Leftrightarrow 
      \cG_{\mbphi}(\mbepsilon \semi \mbx) 
      & \defeq \mbm_{\mbphi}(\mbx) + t \mbepsilon, \quad 
      \mbepsilon \sim \cN(\mbzero, \mbI),
    \end{split} 
  \end{equation*}
  respectively. Thus, expanding out 
  $\cL_{\textsc{nell}}(\mbtheta,\mbphi)$, we have
  \begin{align*}
    \cL_{\textsc{nell}}(\mbtheta,\mbphi) 
    & =
    \E_{q^*(\mbx) q_{\mbphi}(\mbz\g\mbx)} 
    [ - \log p_{\mbtheta}(\mbx\g\mbz) ] \\
    & =
    \E_{q^*(\mbx) p(\mbepsilon)} 
    [ - \log p_{\mbtheta}(\mbx \g \cG_{\mbphi}(\mbepsilon \semi \mbx)) ] \\
    \begin{split}
      & =
      \frac{1}{2\tau^2}
      \E_{q^*(\mbx) p(\mbepsilon)} 
      [ \| \mbx - \mbmu_{\mbtheta}(\cG_{\mbphi}(\mbepsilon \semi \mbx)) \|_2^2 ] \\
      & \qquad + \frac{D}{2} \log 2 \pi \tau^2 
    \end{split} \\
    \begin{split}
      & = \frac{1}{2\tau^2}
      \E_{q^*(\mbx) p(\mbepsilon)} 
      [ \| \mbx - \mbmu_{\mbtheta}(\mbm_{\mbphi}(\mbx) + t \mbepsilon) \|_2^2 ] \\
      & \qquad + \frac{D}{2} \log 2 \pi \tau^2      
    \end{split} \\
    \begin{split}
      & \to \frac{1}{2\tau^2}
      \E_{q^*(\mbx)} 
      [ \| \mbx - \mbmu_{\mbtheta}(\mbm_{\mbphi}(\mbx)) \|_2^2 ] \\
      & \qquad + \frac{D}{2} \log 2 \pi \tau^2 
      , \qquad \textnormal{as } t \to 0
    \end{split} \\
    & = 
    \gamma_1 \ell_{\textsc{const}}^{\textnormal{reverse}}(\mbtheta,\mbphi) + \delta_1 \\
    & \propto \ell_{\textsc{const}}^{\textnormal{reverse}}(\mbtheta,\mbphi)
  \end{align*}
  where $\gamma_1 = \frac{1}{2\tau^2}$ and 
  $\delta_1 = \frac{D}{2} \log \frac{\pi}{\gamma_1}$.

  A similar analysis can be carried out for 
  $\cL_{\textsc{nelp}}(\mbtheta,\mbphi)$ and 
  $\ell_{\textsc{const}}^{\textnormal{forward}}(\mbtheta,\mbphi)$.
\end{proof}

\section{Proof of \Cref{prop:cyclegan_reverse_gan_general}}
\label{sec:proof_of_cyclegan_reverse_gan_general}

\begin{proof}
  To instantiate
  $\cL_{\textsc{gan}}^{\textnormal{reverse}}(\mbalpha \g \mbphi)$ of
  \cref{eq:cyclegan_reverse_gan_general}, it suffices to show that
  $- f_{\textsc{gan}}^*(f_{\textsc{gan}}'(r_{\mbalpha}(\mbz\semi\mbx))) = 
  \log \cD_{\mbalpha}(\mbz\semi\mbx)$ and
  $f_{\textsc{gan}}'(r_{\mbalpha}(\mbz\semi\mbx)) = 
  \log (1 - \cD_{\mbalpha}(\mbz\semi\mbx))$, where 
  $\cD_{\mbalpha}(\mbz \semi \mbx) \defeq 
  1 - \sigma(\log r_{\mbalpha}(\mbz \semi \mbx))$.
First we compute the first derivative $f_{\textsc{gan}}'$ and the 
convex dual $f_{\textsc{gan}}^*$ of $f_{\textsc{gan}}$, which involve 
straightforward calculations,
\begin{align*}
  f_{\textsc{gan}}'(u) &= \log \sigma (\log u), \\
  f_{\textsc{gan}}^*(t) &= - \log (1 - \exp t).
\end{align*}
Thus, the composition $f_{\textsc{gan}}^* \circ f_{\textsc{gan}}': u 
\mapsto f_{\textsc{gan}}^*(f_{\textsc{gan}}'(u))$ can be
simplified as
\begin{align*}
  f_{\textsc{gan}}^*(f_{\textsc{gan}}'(u)) 
  &= - \log (1 - \exp f_{\textsc{gan}}'(u)) \\
  &= - \log (1 - \sigma (\log u)).
\end{align*}
Applying $f_{\textsc{gan}}'$ and $f_{\textsc{gan}}^* \circ f_{\textsc{gan}}'$ 
to $r_{\mbalpha}(\mbz\semi\mbx)$, we have
\begin{align*}
  f_{\textsc{gan}}'(r_{\mbalpha}(\mbz\semi\mbx)) 
  & = \log \sigma (\log r_{\mbalpha}(\mbz\semi\mbx)) \\
  & = \log (1 - \cD_{\mbalpha}(\mbz\semi\mbx)),
\end{align*}
and
\begin{align*}
  f_{\textsc{gan}}^*(f_{\textsc{gan}}'(r_{\mbalpha}(\mbz\semi\mbx)))
  & = - \log (1 - \sigma (\log r_{\mbalpha}(\mbz\semi\mbx))) \\
  & = - \log \cD_{\mbalpha}(\mbz\semi\mbx),
\end{align*}
respectively, as required.
\end{proof}

\section{Proof of \Cref{prop:cyclegan_reverse_gan_reduction}} 
\label{sec:proof_of_cyclegan_reverse_gan_reduction}

\begin{proof}
  Through reparameterization of $q_{\mbphi}(\mbz\g\mbx)$, we have
  \begin{equation*}
    \begin{split}
      \cL_{\textsc{gan}}^{\textnormal{reverse}}(\mbalpha \g \mbphi)
      & =
      \E_{q^*(\mbx) p^*(\mbz)} 
      [ \log \cD_{\mbalpha}(\mbz \semi \mbx) ] \\
      & + \E_{q^*(\mbx) p(\mbepsilon)} 
      [ \log(1 - \cD_{\mbalpha}(\cG_{\mbphi}(\mbepsilon \semi \mbx) \semi \mbx)) ].
    \end{split}
  \end{equation*}
  By specifying a discriminator 
  $\cD_{\mbalpha}(\mbz \semi \mbx) = \mbD_{\mbalpha}(\mbz)$
  which ignores auxiliary input $\mbx$, and mapping 
  $\cG_{\mbphi}(\mbepsilon \semi \mbx) = \mbm_{\mbphi}(\mbx)$
  which ignores noise input $\mbepsilon$, this reduces to 
  \begin{align*}
    \begin{split}
      \cL_{\textsc{gan}}^{\textnormal{reverse}}(\mbalpha \g \mbphi)
      & =
      \E_{p^*(\mbz)} 
      [ \log \mbD_{\mbalpha}(\mbz) ] \\
      & + \E_{q^*(\mbx)} 
      [ \log(1 - \mbD_{\mbalpha}(\mbm_{\mbphi}(\mbx))) ]
    \end{split} \\
    & = \ell_{\textsc{gan}}^{\textnormal{reverse}}(\mbalpha \g \mbphi),
  \end{align*}
  as required.
\end{proof}

\section{Proof of \Cref{prop:dm_loss_kl_equivalence}}
\label{sec:proof_of_dm_loss_kl_equivalence}

\begin{proof}
  Expanding out 
  $\cL_{\textsc{gan}_\textsc{C}}^{\textnormal{reverse}}(\mbphi \g \mbalpha)$,
  we have
  \begin{align*}
    \cL_{\textsc{gan}_\textsc{C}}^{\textnormal{reverse}}(\mbphi \g \mbalpha)
    & = 
    \E_{q^*(\mbx) q_{\mbphi}(\mbz\g\mbx)} 
    \left [ - \log \frac{\cD_{\mbalpha}(\mbz \semi \mbx)}
                        {1 - \cD_{\mbalpha}(\mbz \semi \mbx)} \right ] \\
    & = 
    \E_{q^*(\mbx) q_{\mbphi}(\mbz\g\mbx)} 
    \left [ 
      \log \frac{\sigma(\log r_{\mbalpha}(\mbz \semi \mbx))}
                {1 - \sigma(\log r_{\mbalpha}(\mbz \semi \mbx))}
    \right ] \\
    & = 
    \E_{q^*(\mbx) q_{\mbphi}(\mbz\g\mbx)} 
    [ \log r_{\mbalpha}(\mbz \semi \mbx) ] \\
    & \defeq
    \cL_{\textsc{kl}}^{\textnormal{latent}}(\mbphi \g \mbalpha).
  \end{align*}
  Hence, 
  $\cL_{\textsc{gan}_\textsc{C}}^{\textnormal{reverse}}(\mbphi \g \mbalpha) 
  =\cL_{\textsc{kl}}^{\textnormal{latent}}(\mbphi \g \mbalpha)$ as required.
\end{proof}



\section{Relation to KL Importance Estimation Procedure (KLIEP)}
\label{sec:relation_to_kliep}

We now discuss the connections to \gls{KLIEP} \citep{Sugiyama2008}.
Consider the same problem setting as in 
\cref{sub:approximate_divergence_minimization} 
where we wish to use a parameterized function $r_{\mbalpha}$ to 
estimate the exact density ratio,
\begin{equation*}
  r_{\mbalpha}(\mbz \semi \mbx) 
  \simeq 
  r^*(\mbz \semi \mbx) 
  \defeq 
  \frac{q_{\mbphi}(\mbz\g\mbx)}{p^*(\mbz)}.
\end{equation*}
We can view $r_{\mbalpha}(\mbz \semi \mbx)$ as the correction factor
required for $p^*(\mbz)$ to match $q_{\mbphi}(\mbz\g\mbx)$. 
This gives rise to an estimator of $q_{\mbphi}(\mbz\g\mbx)$,
\begin{equation*}
  q_{\mbalpha}(\mbz\g\mbx)
  \defeq 
  r_{\mbalpha}(\mbz \semi \mbx) p^*(\mbz)
  \simeq  
  q_{\mbphi}(\mbz\g\mbx).
\end{equation*}
Although in our specific problem setting, the density 
$q_{\mbphi}(\mbz\g\mbx)$ is tractable, we nonetheless fit an auxiliary 
model $q_{\mbalpha}(\mbz\g\mbx)$ to it as a means of fitting the 
underlying density ratio estimator $r_{\mbalpha}(\mbz \semi \mbx)$.

In particular, consider \emph{minimizing} the \gls{KL} divergence 
between $q_{\mbphi}(\mbz\g\mbx)$ and $q_{\mbalpha}(\mbz\g\mbx)$ with 
respect to $\mbalpha$,
\begin{align*}
  \begin{split}
    \E_{q^*(\mbx)} & \KL{q_{\mbphi}(\mbz\g\mbx)}{q_{\mbalpha}(\mbz\g\mbx)} \\
    & \defeq
    \E_{q^*(\mbx)}
    \E_{q_{\mbphi}(\mbz\g\mbx)} 
    \left [ 
      \log \frac{q_{\mbphi}(\mbz\g\mbx)}{q_{\mbalpha}(\mbz\g\mbx)} 
    \right ],  
  \end{split} \\
  \begin{split}
    & =
    \E_{q^*(\mbx)}
    \E_{q_{\mbphi}(\mbz\g\mbx)} 
    \left [ 
      \log \frac{q_{\mbphi}(\mbz\g\mbx)}{p^*(\mbz) r_{\mbalpha}(\mbz \semi \mbx)}
    \right ],  
  \end{split} \\
  \begin{split}
    & = -
    \E_{q^*(\mbx)}
    \E_{q_{\mbphi}(\mbz\g\mbx)} 
    [ \log r_{\mbalpha}(\mbz \semi \mbx) ] + \textrm{const}.
  \end{split} \\
\end{align*}
Hence, this is equivalent to \emph{maximizing}
\begin{equation*}
  \E_{q^*(\mbx)}
  \E_{q_{\mbphi}(\mbz\g\mbx)} 
  [ \log r_{\mbalpha}(\mbz \semi \mbx) ].
\end{equation*}
Now, for the conditional $q_{\mbalpha}(\mbz\g\mbx)$ to be a 
probability density function, its integral must sum to one,
\begin{equation*}
  \int q_{\mbalpha}(\mbz\g\mbx) q^*(\mbx) d\mbx d\mbz = 1.
\end{equation*}
Rewriting this integral, we have the constraint
\begin{align*}
  \int q_{\mbalpha}(\mbz\g\mbx) q^*(\mbx) d\mbx d\mbz
  & = \int r_{\mbalpha}(\mbz \semi \mbx) p^*(\mbz) q^*(\mbx) d\mbx d\mbz \\ 
  & = \E_{q^*(\mbx) p^*(\mbz)} [r_{\mbalpha}(\mbz \semi \mbx)] = 1.
\end{align*}
Combined, we have the following constrained optimization problem,
\begin{subequations} 
  \begin{align*}
    \max_{\mbalpha} \qquad & 
    \E_{q^*(\mbx)} \E_{q_{\mbphi}(\mbz\g\mbx)} 
    [ \log r_{\mbalpha}(\mbz \semi \mbx) ] \\
    \textnormal{subject to} \qquad &
    \E_{q^*(\mbx) p^*(\mbz)} [r_{\mbalpha}(\mbz \semi \mbx) - 1] = 0.
  \end{align*}
\end{subequations}
Through the method of Lagrange multipliers, this can be cast as an 
unconstrained optimization problem with objective,
\begin{equation*}
  \begin{split}
    \cL_{\textsc{kliep}}^{\textnormal{latent}}(\mbalpha \g \mbphi)
    & \defeq \E_{q^*(\mbx)} \E_{q_{\mbphi}(\mbz\g\mbx)}
    [ \log r_{\mbalpha}(\mbz \semi \mbx) ] \\ 
    & \qquad
    - \lambda \E_{q^*(\mbx) p^*(\mbz)}
    [r_{\mbalpha}(\mbz \semi \mbx) - 1],
  \end{split}  
\end{equation*}
where $\lambda$ is the Lagrange multiplier. For $\lambda=1$, 
$\cL_{\textsc{kliep}}^{\textnormal{latent}}(\mbalpha \g \mbphi)$ 
trivially reduces to
$\cL_{\textsc{kl}}^{\textnormal{latent}}(\mbalpha \g \mbphi)$.

\section{Summary of Definitions}
\label{sec:summary_of_definitions}

In this section, we summarize the definitions of the losses defined 
in the proposed \gls{VI} framework of 
\cref{sec:variational_inference_with_implicit_priors,sec:symmetric_joint_matching_variational_inference}, and underscore 
the relationships to their respective counterparts in the 
\gls{CYCLEGAN} framework of \cref{sec:deriving_cyclegan_as_a_special_case}.

\begin{table*}[t]
\caption{Relevant latent and observed space $f$-divergences 
         instantiated for particular settings of $f$.}
\label{tab:f_divergences}
\vskip 0.15in
\begin{center}
\begin{small}
\begin{sc}
\def\arraystretch{2}
\begin{tabular}{lcccc}
\toprule
& & Reverse \gls{KL} & \gls{GAN} \\
& $f(u)$ & $\displaystyle u \log u$ & $\displaystyle u \log u - (u + 1) \log (u + 1)$ \\
\midrule
Latent   & $\displaystyle \E_{q^*(\mbx)} \Df{p^*(\mbz)}{q_{\mbphi}(\mbz\g\mbx)}$ & 
         $\displaystyle \E_{q^*(\mbx)} \KL{q_{\mbphi}(\mbz\g\mbx)}{p^*(\mbz)}$ &
         $\displaystyle 2 \cdot \E_{q^*(\mbx)} \JS{p^*(\mbz)}{q_{\mbphi}(\mbz\g\mbx)} - \log 4$ \\
Observed & $\displaystyle \E_{p^*(\mbz)} \Df{q^*(\mbx)}{p_{\mbtheta}(\mbx\g\mbz)}$ &
         $\displaystyle \E_{p^*(\mbz)} \KL{p_{\mbtheta}(\mbx\g\mbz)}{q^*(\mbx)}$ & 
         $\displaystyle 2 \cdot \E_{p^*(\mbz)} \JS{q^*(\mbx)}{p_{\mbtheta}(\mbx\g\mbz)} - \log 4$ \\
\bottomrule
\end{tabular}
\end{sc}
\end{small}
\end{center}
\vskip -0.1in
\end{table*}

\Cref{tab:f_divergences} summarizes the settings of convex function
$f: \bbR_{+} \to \bbR$ that recover the reverse \gls{KL} divergence
terms within the \gls{ELBO} and \gls{APLBO}, and the \gls{JS} 
divergence (up to constants) that \glspl{GAN} are known to minimize.

\begin{table}[t]
\caption{Calculations for convex functions.}
\label{tab:f_calculations}
\vskip 0.15in
\begin{center}
\begin{small}
\begin{sc}
\def\arraystretch{2}
\begin{tabular}{lcc}
\toprule
& Reverse \gls{KL} & \gls{GAN} \\
$\displaystyle f(u)$ & $\displaystyle u \log u$ & $\displaystyle u \log u - (u + 1) \log (u + 1)$ \\
\midrule
$\displaystyle f^{\star}(t)$ & $\displaystyle \exp (t - 1)$ & $\displaystyle - \log (1 - \exp t)$ \\
$\displaystyle f'(u)$ & $\displaystyle 1 + \log u$ & $\displaystyle \log \sigma (\log u)$ \\
$\displaystyle f^{\star}(f'(u))$ & $\displaystyle u$ & $\displaystyle - \log (1 - \sigma (\log u))$ \\
\bottomrule
\end{tabular}
\end{sc}
\end{small}
\end{center}
\vskip -0.1in
\end{table}

\Cref{tab:f_calculations} gives the calculations of the terms 
necessary to explicitly write down instances of the generalized 
variational lower bound for particular convex functions $f$---namely 
the convex dual $f^{\star}$, the first derivative $f'$ and the 
composition $f^{\star} \circ f'$.

\Cref{tab:f_variational_lower_bounds} gives instances of the 
variational lower bound that approximate the latent and observed
space \gls{KL} divergences within the \gls{ELBO} and \gls{APLBO}, 
respectively. Additionally, it gives generalized \emph{stochastic} 
formulations of the \gls{GAN} objectives in the \gls{CYCLEGAN} 
framework, while \cref{tab:gan_stochastic_deterministic} lists their 
\emph{deterministic} counterpart.

Lastly, \cref{tab:elcp_cycle_consistency} gives forward and reverse 
cycle-consistency constraints in the \gls{CYCLEGAN} framework, and 
the specific class of Gaussian likelihoods and posteriors that 
instantiates these constraints (in the limit).

\begin{sidewaystable*}[t]
\caption{Instances of variational lower bounds on the relevant latent 
         and observed space $f$-divergences.}
\label{tab:f_variational_lower_bounds}
\vskip 0.15in
\begin{center}
\begin{small}
\begin{sc}
\def\arraystretch{4}
\begin{tabular}{lccc}
\toprule
& & Reverse \gls{KL} & \gls{GAN} \\ 
& $f(u)$ & $\displaystyle u \log u$ & $\displaystyle u \log u - (u + 1) \log (u + 1)$ \\
\midrule
Latent   & 
$\begin{aligned}
\cL_f^{\textnormal{latent}}(\mbalpha \g \mbphi)
& \defeq 
\E_{q^*(\mbx) q_{\mbphi}(\mbz\g\mbx)} 
[ f'(r_{\mbalpha}(\mbz \semi \mbx)) ] \\
& \quad - 
\E_{q^*(\mbx) p^*(\mbz)} 
[ f^{\star}(f'(r_{\mbalpha}(\mbz \semi \mbx))) ]
\end{aligned}$ & 
$\begin{aligned}
  \cL_{\textsc{kl}}^{\textnormal{latent}}(\mbalpha \g \mbphi)
  & \defeq 
  \E_{q^*(\mbx) q_{\mbphi}(\mbz\g\mbx)} 
  [ \log r_{\mbalpha}(\mbz \semi \mbx) ] \\
  & \quad
  - \E_{q^*(\mbx) p^*(\mbz)} [ 
    r_{\mbalpha}(\mbz \semi \mbx) - 1
  ]
\end{aligned}$ &
$\begin{aligned}
  \cL_{\textsc{gan}}^{\textnormal{reverse}}(\mbalpha \g \mbphi)
  & \defeq
  \E_{q^*(\mbx) q_{\mbphi}(\mbz\g\mbx)} 
  [ \log \sigma(\log r_{\mbalpha}(\mbz \semi \mbx)) ] \\
  & + 
  \E_{q^*(\mbx) p^*(\mbz)} 
  [ \log (1 - \sigma(\log r_{\mbalpha}(\mbz \semi \mbx))) ] 
\end{aligned}$ 
\\
Observed & 
$\begin{aligned}
  \cL_f^{\textnormal{observed}}(\mbbeta \g \mbtheta)
  & \defeq 
  \E_{p^*(\mbz) p_{\mbtheta}(\mbx\g\mbz)} 
  [ f'(r_{\mbbeta}(\mbx \semi \mbz)) ] \\
  & \quad
  - \E_{p^*(\mbz) q^*(\mbx)} 
  [ f^{\star}(f'(r_{\mbbeta}(\mbx \semi \mbz))) ]
\end{aligned}$ &
$\begin{aligned}
  \cL_{\textsc{kl}}^{\textnormal{observed}}(\mbbeta \g \mbtheta)
  & \defeq 
  \E_{p^*(\mbz) p_{\mbtheta}(\mbx\g\mbz)} 
  [ \log r_{\mbbeta}(\mbx \semi \mbz) ] \\
  & \quad
  - \E_{p^*(\mbz) q^*(\mbx)} [
    r_{\mbbeta}(\mbx \semi \mbz) - 1
  ]
\end{aligned}$ & 
$\begin{aligned}
  \cL_{\textsc{gan}}^{\textnormal{forward}}(\mbbeta \g \mbtheta)
  & \defeq
  \E_{p^*(\mbz) p_{\mbtheta}(\mbx\g\mbz)} 
  [ \log \sigma(\log r_{\mbbeta}(\mbx \semi \mbz)) ] \\
  & + 
  \E_{p^*(\mbz) q^*(\mbx)} 
  [ \log (1 - \sigma(\log r_{\mbbeta}(\mbx \semi \mbz))) ] 
\end{aligned}$ 
\\
\bottomrule
\end{tabular}
\end{sc}
\end{small}
\end{center}
\vskip -0.1in
\end{sidewaystable*}

\begin{table*}[t]
\caption{General stochastic \gls{GAN} objectives and their 
         deterministic counterparts.}
\label{tab:gan_stochastic_deterministic}
\vskip 0.15in
\begin{center}
\begin{small}
\begin{sc}
\def\arraystretch{3}
\begin{tabular}{lcc}
\toprule
& Stochastic & Deterministic \\
\midrule
reverse &
$\begin{aligned}
  \cL_{\textsc{gan}}^{\textnormal{reverse}}(\mbalpha \g \mbphi)
  & \defeq
  \E_{q^*(\mbx) p^*(\mbz)} 
  [ \log \cD_{\mbalpha}(\mbz \semi \mbx) ] \\
  & + 
  \E_{q^*(\mbx) p(\mbepsilon)} 
  [ \log (1 - \cD_{\mbalpha}(\cG_{\mbphi}(\mbepsilon \semi \mbx) \semi \mbx)) ]
\end{aligned}$ &
$\begin{aligned}
  \ell_{\textsc{gan}}^{\textnormal{reverse}}(\mbalpha \g \mbphi)
  & \defeq 
  \E_{p^*(\mbz)} 
  [ \log \mbD_{\mbalpha}(\mbz) ] \\
  & \quad + \E_{q^*(\mbx)} 
  [ \log(1 - \mbD_{\mbalpha}(\mbm_{\mbphi}(\mbx))) ]
\end{aligned}$
\\
forward &
$\begin{aligned}
  \cL_{\textsc{gan}}^{\textnormal{forward}}(\mbbeta \g \mbtheta)
  & \defeq
  \E_{p^*(\mbz) q^*(\mbx)} 
  [ \log \cD_{\mbbeta}(\mbx \semi \mbz) ] \\
  & + 
  \E_{p^*(\mbz) p(\mbxi)} 
  [ \log (1 - \cD_{\mbbeta}(\cF_{\mbtheta}(\mbxi \semi \mbz) \semi \mbz)) ]
\end{aligned}$ &
$\begin{aligned}
  \ell_{\textsc{gan}}^{\textnormal{forward}}(\mbbeta \g \mbtheta)
  & \defeq 
  \E_{p^*(\mbx)} 
  [ \log \mbD_{\mbbeta}(\mbx) ] \\
  & \quad + \E_{p^*(\mbz)} 
  [ \log (1 - \mbD_{\mbbeta}(\mbmu_{\mbtheta}(\mbz))) ]
\end{aligned}$ \\
\bottomrule
\end{tabular}
\end{sc}
\end{small}
\end{center}
\vskip -0.1in
\end{table*}


\begin{table*}[t]
\caption{Negative expected log conditionals and the cycle-consistency constraints.}
\label{tab:elcp_cycle_consistency}
\vskip 0.15in
\begin{center}
\begin{small}
\begin{sc}
\def\arraystretch{2}
\begin{tabular}{cccc}
\toprule
\multicolumn{2}{c}{Gaussian} & \multicolumn{2}{c}{Degenerate} \\
$p_{\mbtheta}(\mbx\g\mbz)$ & $q_{\mbphi}(\mbz\g\mbx)$ & 
$p_{\mbtheta}(\mbx\g\mbz)$ & $q_{\mbphi}(\mbz\g\mbx)$ \\
$\cN(\mbx \g \mbmu_{\mbtheta}(\mbz), \tau^2 \mbI)$ &
$\cN(\mbz \g \mbm_{\mbphi}(\mbx), t^2 \mbI)$ &
$\to \delta(\mbx - \mbmu_{\mbtheta}(\mbz))$ &
$\to \delta(\mbz - \mbm_{\mbphi}(\mbx))$ \\
\midrule
\multicolumn{2}{c}{$\begin{aligned}
  \cL_{\textsc{nell}}(\mbtheta,\mbphi) \defeq
  \frac{1}{2\tau^2}
  \E_{q^*(\mbx) p(\mbepsilon)} 
  [ \| \mbx - \mbmu_{\mbtheta}(\mbm_{\mbphi}(\mbx) + t \mbepsilon) \|_2^2 ] 
  + \frac{D}{2} \log 2 \pi \tau^2      
\end{aligned}$} &
\multicolumn{2}{c}{$\ell_{\textsc{const}}^{\textnormal{reverse}}(\mbtheta,\mbphi) \defeq
\E_{q^*(\mbx)} 
[ \| \mbx - \mbmu_{\mbtheta}(\mbm_{\mbphi}(\mbx)) \|_2^2 ]$} \\
\multicolumn{2}{c}{$\begin{aligned}
  \cL_{\textsc{nelp}}(\mbtheta,\mbphi) \defeq
  \frac{1}{2t^2}
  \E_{p^*(\mbz) p(\mbxi)} 
  [ \| \mbz - \mbm_{\mbphi}(\mbmu_{\mbtheta}(\mbz) + \tau \mbxi) \|_2^2 ] 
  + \frac{K}{2} \log 2 \pi t^2      
\end{aligned}$} &
\multicolumn{2}{c}{
$\ell_{\textsc{const}}^{\textnormal{forward}}(\mbtheta,\mbphi) \defeq 
\E_{p^*(\mbz)} 
[ \| \mbz - \mbm_{\mbphi}(\mbmu_{\mbtheta}(\mbz)) \|_2^2 ]$} \\
\bottomrule
\end{tabular}
\end{sc}
\end{small}
\end{center}
\vskip -0.1in
\end{table*}

\end{document}

%% file: preamble/preamble.tex
\usepackage[T1]{fontenc}
\usepackage[utf8]{inputenc}
\usepackage{tgtermes}

\usepackage{amssymb}
\newcommand{\mathbold}[1]{\ensuremath{\boldsymbol{\mathbf{#1}}}}

\usepackage{scalefnt,letltxmacro}
\LetLtxMacro{\oldtextsc}{\textsc}
\renewcommand{\textsc}[1]{\oldtextsc{\scalefont{1.10}#1}}
\usepackage[ttdefault=true]{AnonymousPro}

\usepackage{mathtools}
\usepackage{amsfonts}
\usepackage{amsmath}
\usepackage{amsthm}

\newtheorem{theorem}{Theorem}
\newtheorem{corollary}{Corollary}[theorem]

\newtheorem{proposition}{Proposition}
\newtheorem{lemma}{Lemma}

\theoremstyle{definition}

\usepackage{empheq}

\usepackage[english]{babel}
\usepackage[parfill]{parskip}
\usepackage{afterpage}
\usepackage{enumitem}
\usepackage{framed}
\usepackage{setspace}

\usepackage{lineno}

\usepackage{ragged2e}

\newcounter{parcount}

\DeclareRobustCommand{\parhead}[1]{\textbf{#1}~}

\usepackage[usenames,dvipsnames]{xcolor}
\definecolor{shadecolor}{gray}{0.9}

\usepackage{graphicx}
\usepackage[labelfont=bf]{caption}
\usepackage[format=hang]{subcaption}
\usepackage{wrapfig}

\usepackage{booktabs}
\usepackage{array}

\usepackage[sort&compress]{natbib}
\usepackage{bibentry}

\nobibliography*

\usepackage[colorlinks,linktoc=all]{hyperref}
\usepackage[all]{hypcap}
\hypersetup{citecolor=MidnightBlue}
\hypersetup{linkcolor=MidnightBlue}
\hypersetup{urlcolor=MidnightBlue}

\usepackage[acronym,smallcaps,nowarn]{glossaries}
\glsdisablehyper
\makeglossaries

\usepackage[nameinlink]{cleveref}
\crefname{section}{\S}{\S\S}
\Crefname{section}{\S}{\S\S}
\creflabelformat{equation}{#2\textup{#1}#3}

\usepackage[colorinlistoftodos,
            disable,
            textsize=tiny,
            linecolor=red!30,
            bordercolor=red!30,
            backgroundcolor=red!10]{todonotes}

\usepackage[most]{tcolorbox}

\newtcolorbox{incomplete}{skin=enhancedmiddle jigsaw,breakable,
  parbox=false,boxrule=0mm,leftrule=2mm,boxsep=0mm,
  arc=0mm,outer arc=0mm,left=3mm,right=3mm,top=1mm,bottom=1mm,
  toptitle=1mm,bottomtitle=1mm,oversize,colback=yellow!5!white,
  colframe=yellow}

%% file: preamble/preamble_acronyms.tex
\newacronym{ELL}{ell}{expected log likelihood}
\newacronym{ELP}{elp}{expected log posterior}
\newacronym{EM}{em}{expectation maximization}
\newacronym{AVB}{avb}{Adversarial Variational Bayes}
\newacronym{AAE}{aae}{adversarial autoencoder}

\newacronym{LVM}{lvm}{latent variable model}
\newacronym{ILVM}{ilvm}{implicit latent variable model}

\newacronym{MAP}{map}{maximum a posteriori}

\newacronym{GAN}{gan}{generative adversarial network}
\newacronym{CYCLEGAN}{cycle\-gan}{cycle-consistent adversarial learning}
\newacronym{VAE}{vae}{variational autoencoder}

\newacronym{KL}{kl}{Kullback-Leibler}
\newacronym{JS}{js}{Jensen-Shannon}

\newacronym{ELBO}{elbo}{evidence lower bound}
\newacronym{APLBO}{aplbo}{aggregate posterior lower bound}

\newacronym{MC}{mc}{Monte Carlo}
\newacronym{MCMC}{mcmc}{Markov chain Monte Carlo}

\newacronym{VI}{vi}{variational inference}
\newacronym{LFVI}{lfvi}{likelihood-free variational inference}

\newacronym{KLIEP}{kliep}{KL importance estimation procedure}
\newacronym{LSIF}{lsif}{least squares importance fitting}

\newacronym{PPCA}{ppca}{probabilistic principal component analysis}
\newacronym{PCA}{pca}{principal component analysis}
\newacronym{FA}{fa}{factor analysis}

\newacronym{DLGM}{dlgm}{deep latent Gaussian model}
\newacronym{MLP}{mlp}{multilayer perceptron}

\newacronym{DGM}{dgm}{directed graphical model}

\newacronym{ABC}{abc}{approximate Bayesian computation}

\newacronym{BIGAN}{bigan}{Bidirectional GAN}
\newacronym{ALI}{ali}{adversarially learned inference}

\newacronym{MSE}{mse}{mean-squared error}

\newacronym{DRE}{dre}{density ratio estimation}

\newacronym{LSGAN}{lsgan}{Least-Squares \textsc{gan}}

\newacronym{DM}{dm}{divergence minimization}

\newacronym{HIM}{him}{hierarchical implicit model}

%% file: preamble/preamble_math.tex
\newcommand{\g}{\,|\,}

\newcommand{\semi}{;}
\newcommand{\cd}{\,\cdot\,}

\newcommand{\nestedmathbold}[1]{{\mathbold{#1}}}

\newcommand{\mbb}{\nestedmathbold{b}}

\newcommand{\mbm}{\nestedmathbold{m}}

\newcommand{\mbu}{\nestedmathbold{u}}

\newcommand{\mbx}{\nestedmathbold{x}}

\newcommand{\mbz}{\nestedmathbold{z}}

\newcommand{\mbD}{\nestedmathbold{D}}

\newcommand{\mbI}{\nestedmathbold{I}}

\newcommand{\mbW}{\nestedmathbold{W}}
\newcommand{\mbX}{\nestedmathbold{X}}

\newcommand{\mbZ}{\nestedmathbold{Z}}

\newcommand{\mbalpha}{\nestedmathbold{\alpha}}
\newcommand{\mbbeta}{\nestedmathbold{\beta}}

\newcommand{\mbepsilon}{\nestedmathbold{\epsilon}}

\newcommand{\mblambda}{\nestedmathbold{\lambda}}
\newcommand{\mbmu}{\nestedmathbold{\mu}}

\newcommand{\mbomega}{\nestedmathbold{\omega}}
\newcommand{\mbphi}{\nestedmathbold{\phi}}

\newcommand{\mbtheta}{\nestedmathbold{\theta}}

\newcommand{\mbxi}{\nestedmathbold{\xi}}

\newcommand{\mbPsi}{\nestedmathbold{\Psi}}
\newcommand{\mbSigma}{\nestedmathbold{\Sigma}}

\newcommand{\mbzero}{\nestedmathbold{0}}

\DeclareRobustCommand{\KL}[2]{\ensuremath{\textsc{kl}\left[#1\;\|\;#2\right]}}
\DeclareRobustCommand{\JS}[2]{\ensuremath{\textsc{js}\left[#1\;\|\;#2\right]}}
\DeclareRobustCommand{\KLSYMM}[2]{\ensuremath{\textsc{kl}_{\textsc{symm}}\left[#1\;\|\;#2\right]}}
\DeclareRobustCommand{\Df}[2]{\ensuremath{\mathcal{D}_f\left[#1\;\|\;#2\right]}}

\newcommand{\diag}{\textrm{diag}}

\newcommand{\cD}{\mathcal{D}}

\newcommand{\cF}{\mathcal{F}}
\newcommand{\cG}{\mathcal{G}}

\newcommand{\cL}{\mathcal{L}}
\newcommand{\cN}{\mathcal{N}}

\newcommand{\cQ}{\mathcal{Q}}
\newcommand{\cR}{\mathcal{R}}

\newcommand{\E}{\mathbb{E}}
\newcommand{\bbH}{\mathbb{H}}
\newcommand{\bbR}{\mathbb{R}}

\newcommand{\defeq}{\coloneqq}



%% file: preamble/preamble_tikz.tex
\usepackage{tikz}

\usetikzlibrary{bayesnet} 

\pgfdeclarelayer{edgelayer}
\pgfdeclarelayer{nodelayer}
\pgfsetlayers{edgelayer,nodelayer,main}

\definecolor{hexcolor0xbfbfbf}{rgb}{0.749,0.749,0.749}

\tikzset{>=latex}
\tikzstyle{none}   = [inner sep=0pt]
\tikzstyle{line}   = [ -, thick, shorten <=1pt, shorten >=1pt ]
\tikzstyle{arrow}  = [ ->, thick, shorten <=1pt, shorten >=1pt ]
\tikzstyle{ardash} = [ dashed, ->, thick, shorten <=1pt, shorten >=1pt ]

\tikzstyle{empty}=[circle,opacity=0.0,text opacity=1.0,inner sep=0pt]
\tikzstyle{box}=[rectangle,fill=White,draw=Black]
\tikzstyle{filled}=[circle,thick,fill=hexcolor0xbfbfbf,draw=Black]
\tikzstyle{hollow}=[circle,thick,fill=White,draw=Black]
\tikzstyle{param}=[rectangle,fill=Black,draw=Black,inner sep=0pt,minimum width=4pt,minimum height=4pt]
\tikzstyle{paramhollow}=[rectangle,thick,fill=White,draw=Black,inner sep=0pt,minimum
width=4pt,minimum height=4pt]

\usepackage{pgfplots}                               
\pgfplotsset{compat=newest}
\pgfplotsset{plot coordinates/math parser=false}
\newlength\figureheight
\newlength\figurewidth
\newlength\figureheightsmall
\newlength\figurewidthsmall
\definecolor{POSTcolor}{rgb}{0.48, 0.20, 0.58} 
\definecolor{Qcolor}{rgb}{0.00, 0.53, 0.22} 

%% file: main.bbl
\begin{thebibliography}{39}
\providecommand{\natexlab}[1]{#1}
\providecommand{\url}[1]{\texttt{#1}}
\expandafter\ifx\csname urlstyle\endcsname\relax
  \providecommand{\doi}[1]{doi: #1}\else
  \providecommand{\doi}{doi: \begingroup \urlstyle{rm}\Url}\fi

\bibitem[Ali \& Silvey(1966)Ali and Silvey]{Ali1966}
Ali, S.~M. and Silvey, S.~D.
\newblock {A General Class of Coefficients of Divergence of One Distribution
  from Another}, 1966.

\bibitem[Bartholomew et~al.(2011)Bartholomew, Knott, and
  Moustaki]{bartholomew2011latent}
Bartholomew, David~J, Knott, Martin, and Moustaki, Irini.
\newblock \emph{{Latent variable models and factor analysis: A unified
  approach}}, volume 904.
\newblock John Wiley {\&} Sons, 2011.

\bibitem[Blei et~al.(2017)Blei, Kucukelbir, and
  McAuliffe]{doi:10.1080/01621459.2017.1285773}
Blei, David~M, Kucukelbir, Alp, and McAuliffe, Jon~D.
\newblock {Variational Inference: A Review for Statisticians}.
\newblock \emph{Journal of the American Statistical Association}, 112\penalty0
  (518):\penalty0 859--877, 2017.
\newblock \doi{10.1080/01621459.2017.1285773}.

\bibitem[Chen et~al.(2016)Chen, Duan, Houthooft, Schulman, Sutskever, and
  Abbeel]{Chen2016}
Chen, Xi, Duan, Yan, Houthooft, Rein, Schulman, John, Sutskever, Ilya, and
  Abbeel, Pieter.
\newblock {InfoGAN: Interpretable Representation Learning by Information
  Maximizing Generative Adversarial Nets}.
\newblock jun 2016.

\bibitem[Ciszar(1967)]{ciszar1967information}
Ciszar, I.
\newblock {Information-type measures of difference of probability distributions
  and indirect observations}.
\newblock \emph{Studia Sci. Math. Hungar.}, 2:\penalty0 299--318, 1967.

\bibitem[Dayan et~al.(1995)Dayan, Hinton, Neal, and
  Zemel]{DBLP:journals/neco/DayanHNZ95}
Dayan, Peter, Hinton, Geoffrey~E, Neal, Radford~M, and Zemel, Richard~S.
\newblock {The Helmholtz machine}.
\newblock \emph{Neural Computation}, 7\penalty0 (5):\penalty0 889--904, 1995.
\newblock \doi{10.1162/neco.1995.7.5.889}.

\bibitem[Donahue et~al.(2016)Donahue, Kr{\"{a}}henb{\"{u}}hl, and
  Darrell]{Donahue2016}
Donahue, Jeff, Kr{\"{a}}henb{\"{u}}hl, Philipp, and Darrell, Trevor.
\newblock {Adversarial Feature Learning}.
\newblock may 2016.

\bibitem[Dumoulin et~al.(2016)Dumoulin, Belghazi, Poole, Mastropietro, Lamb,
  Arjovsky, and Courville]{Dumoulin2016}
Dumoulin, Vincent, Belghazi, Ishmael, Poole, Ben, Mastropietro, Olivier, Lamb,
  Alex, Arjovsky, Martin, and Courville, Aaron.
\newblock {Adversarially Learned Inference}.
\newblock jun 2016.

\bibitem[Frey \& Hinton(1999)Frey and Hinton]{frey1999variational}
Frey, Brendan~J and Hinton, Geoffrey~E.
\newblock {Variational learning in nonlinear Gaussian belief networks}.
\newblock \emph{Neural Computation}, 11\penalty0 (1):\penalty0 193--213, 1999.

\bibitem[Gal \& Ghahramani(2015)Gal and Ghahramani]{Gal2015}
Gal, Yarin and Ghahramani, Zoubin.
\newblock {Dropout as a Bayesian Approximation: Appendix}.
\newblock jun 2015.

\bibitem[Gershman \& Goodman(2014)Gershman and Goodman]{gershman2014amortized}
Gershman, Samuel and Goodman, Noah.
\newblock {Amortized inference in probabilistic reasoning}.
\newblock In \emph{Proceedings of the Annual Meeting of the Cognitive Science
  Society}, volume~36, 2014.

\bibitem[Gneiting \& Raftery(2007)Gneiting and Raftery]{10.2307/27639845}
Gneiting, Tilmann and Raftery, Adrian~E.
\newblock {Strictly Proper Scoring Rules, Prediction, and Estimation}.
\newblock \emph{Journal of the American Statistical Association}, 102\penalty0
  (477):\penalty0 359--378, 2007.
\newblock ISSN 01621459.
\newblock \doi{10.2307/27639845}.

\bibitem[Goodfellow et~al.(2014)Goodfellow, Pouget-Abadie, Mirza, Xu,
  Warde-Farley, Ozair, Courville, and Bengio]{NIPS2014_5423}
Goodfellow, Ian~J., Pouget-Abadie, Jean, Mirza, Mehdi, Xu, Bing, Warde-Farley,
  David, Ozair, Sherjil, Courville, Aaron, and Bengio, Yoshua.
\newblock {Generative Adversarial Networks}.
\newblock In Ghahramani, Z, Welling, M, Cortes, C, Lawrence, N~D, and
  Weinberger, K~Q (eds.), \emph{Advances in Neural Information Processing
  Systems 27}, pp.\  2672--2680. Curran Associates, Inc., jun 2014.

\bibitem[Haario et~al.(1999)Haario, Saksman, and Tamminen]{Haario1999}
Haario, Heikki, Saksman, Eero, and Tamminen, Johanna.
\newblock {Adaptive proposal distribution for random walk Metropolis
  algorithm}.
\newblock \emph{Computational Statistics}, 14\penalty0 (3):\penalty0 375, 1999.
\newblock ISSN 09434062.
\newblock \doi{10.1007/s001800050022}.

\bibitem[Hu et~al.(2017)Hu, Yang, Salakhutdinov, and Xing]{Hu2017}
Hu, Zhiting, Yang, Zichao, Salakhutdinov, Ruslan, and Xing, Eric~P.
\newblock {On Unifying Deep Generative Models}.
\newblock jun 2017.

\bibitem[Husz{\'{a}}r(2017)]{Huszar2017}
Husz{\'{a}}r, Ferenc.
\newblock {Variational Inference using Implicit Distributions}.
\newblock feb 2017.

\bibitem[Jaynes(1968)]{Jaynes1968}
Jaynes, Edwin.
\newblock {Prior Probabilities}.
\newblock \emph{IEEE Transactions on Systems Science and Cybernetics},
  4\penalty0 (3):\penalty0 227--241, 1968.
\newblock ISSN 0536-1567.
\newblock \doi{10.1109/TSSC.1968.300117}.

\bibitem[Jordan et~al.(1999)Jordan, Ghahramani, Jaakkola, and Saul]{Jordan1999}
Jordan, Michael~I., Ghahramani, Zoubin, Jaakkola, Tommi~S., and Saul,
  Lawrence~K.
\newblock {An Introduction to Variational Methods for Graphical Models}.
\newblock \emph{Machine Learning}, 37\penalty0 (2):\penalty0 183--233, 1999.
\newblock ISSN 08856125.
\newblock \doi{10.1023/A:1007665907178}.

\bibitem[Kim et~al.(2017)Kim, Cha, Kim, Lee, and Kim]{pmlr-v70-kim17a}
Kim, Taeksoo, Cha, Moonsu, Kim, Hyunsoo, Lee, Jung~Kwon, and Kim, Jiwon.
\newblock {Learning to Discover Cross-Domain Relations with Generative
  Adversarial Networks}.
\newblock In \emph{Proceedings of the 34th International Conference on Machine
  Learning (ICML)}, volume~70, pp.\  1857--1865, 2017.

\bibitem[Kingma \& Welling(2014)Kingma and Welling]{Kingma2013}
Kingma, Diederik~P and Welling, Max.
\newblock {Auto-Encoding Variational Bayes}.
\newblock In \emph{Proceedings of the 2nd International Conference on Learning
  Representations (ICLR) 2014}, Dec 2014.

\bibitem[Lappalainen \& Honkela(2000)Lappalainen and
  Honkela]{lappalainen2000bayesian}
Lappalainen, Harri and Honkela, Antti.
\newblock {Bayesian non-linear independent component analysis by multi-layer
  perceptrons}.
\newblock In \emph{Advances in independent component analysis}, pp.\  93--121.
  Springer, 2000.

\bibitem[Makhzani et~al.(2015)Makhzani, Shlens, Jaitly, Goodfellow, and
  Frey]{Makhzani2015}
Makhzani, Alireza, Shlens, Jonathon, Jaitly, Navdeep, Goodfellow, Ian, and
  Frey, Brendan.
\newblock {Adversarial Autoencoders}.
\newblock nov 2015.

\bibitem[Mao et~al.(2016)Mao, Li, Xie, Lau, Wang, and Smolley]{Mao2016}
Mao, Xudong, Li, Qing, Xie, Haoran, Lau, Raymond Y.~K., Wang, Zhen, and
  Smolley, Stephen~Paul.
\newblock {Least Squares Generative Adversarial Networks}.
\newblock nov 2016.

\bibitem[Mescheder et~al.(2017)Mescheder, Nowozin, and
  Geiger]{pmlr-v70-mescheder17a}
Mescheder, Lars, Nowozin, Sebastian, and Geiger, Andreas.
\newblock {Adversarial Variational Bayes: Unifying Variational Autoencoders and
  Generative Adversarial Networks}.
\newblock In \emph{Proceedings of the 34th International Conference on Machine
  Learning (ICML)}, volume~70, pp.\  2391--2400, Jan 2017.

\bibitem[Mohamed \& Lakshminarayanan(2017)Mohamed and
  Lakshminarayanan]{Mohamed2016}
Mohamed, Shakir and Lakshminarayanan, Balaji.
\newblock {Learning in Implicit Generative Models}.
\newblock In \emph{The 5th International Conference on Learning
  Representations}, oct 2017.

\bibitem[Nguyen et~al.(2010)Nguyen, Wainwright, and
  Jordan]{DBLP:journals/tit/NguyenWJ10}
Nguyen, XuanLong, Wainwright, Martin~J, and Jordan, Michael~I.
\newblock {Estimating Divergence Functionals and the Likelihood Ratio by Convex
  Risk Minimization}.
\newblock \emph{{IEEE} Trans. Information Theory}, 56\penalty0 (11):\penalty0
  5847--5861, 2010.
\newblock \doi{10.1109/TIT.2010.2068870}.

\bibitem[Nowozin et~al.(2016)Nowozin, Cseke, and Tomioka]{NIPS2016_6066}
Nowozin, Sebastian, Cseke, Botond, and Tomioka, Ryota.
\newblock {f-GAN: Training Generative Neural Samplers using Variational
  Divergence Minimization}.
\newblock In Lee, D~D, Sugiyama, M, Luxburg, U~V, Guyon, I, and Garnett, R
  (eds.), \emph{Advances in Neural Information Processing Systems 29}, pp.\
  271--279. Curran Associates, Inc., jun 2016.

\bibitem[Pu et~al.(2017)Pu, Wang, Henao, Chen, Gan, Li, and
  Carin]{NIPS2017_7020}
Pu, Yuchen, Wang, Weiyao, Henao, Ricardo, Chen, Liqun, Gan, Zhe, Li, Chunyuan,
  and Carin, Lawrence.
\newblock {Adversarial Symmetric Variational Autoencoder}.
\newblock In Guyon, I, Luxburg, U~V, Bengio, S, Wallach, H, Fergus, R,
  Vishwanathan, S, and Garnett, R (eds.), \emph{Advances in Neural Information
  Processing Systems 30}, pp.\  4330--4339. Curran Associates, Inc., 2017.

\bibitem[Rezende et~al.(2014)Rezende, Mohamed, and
  Wierstra]{pmlr-v32-rezende14}
Rezende, Danilo~Jimenez, Mohamed, Shakir, and Wierstra, Daan.
\newblock {Stochastic backpropagation and approximate inference in deep
  generative models}.
\newblock In Xing, Eric~P and Jebara, Tony (eds.), \emph{Proceedings of The
  31st {\ldots}}, volume~32 of \emph{Proceedings of Machine Learning Research},
  pp.\  1278--1286, Bejing, China, jan 2014. PMLR.
\newblock ISBN 9781634393973.
\newblock \doi{10.1051/0004-6361/201527329}.

\bibitem[S{\o}nderby et~al.(2016)S{\o}nderby, Caballero, Theis, Shi, and
  Husz{\'{a}}r]{sonderby2016amortised}
S{\o}nderby, Casper~Kaae, Caballero, Jose, Theis, Lucas, Shi, Wenzhe, and
  Husz{\'{a}}r, Ferenc.
\newblock {Amortised map inference for image super-resolution}.
\newblock \emph{arXiv preprint arXiv:1610.04490}, oct 2016.

\bibitem[Srivastava et~al.(2017)Srivastava, Valkoz, Russell, Gutmann, Sutton,
  Valkov, Russell, Gutmann, and Sutton]{srivastava2017veegan}
Srivastava, Akash, Valkoz, Lazar, Russell, Chris, Gutmann, Michael~U., Sutton,
  Charles, Valkov, Lazar, Russell, Chris, Gutmann, Michael~U., and Sutton,
  Charles.
\newblock {Veegan: Reducing mode collapse in gans using implicit variational
  learning}.
\newblock In \emph{Advances in Neural Information Processing Systems}, pp.\
  3310--3320, may 2017.

\bibitem[Sugiyama et~al.(2008)Sugiyama, Suzuki, Nakajima, Kashima, von
  B{\"{u}}nau, and Kawanabe]{Sugiyama2008}
Sugiyama, Masashi, Suzuki, Taiji, Nakajima, Shinichi, Kashima, Hisashi, von
  B{\"{u}}nau, Paul, and Kawanabe, Motoaki.
\newblock {Direct importance estimation for covariate shift adaptation}.
\newblock \emph{Annals of the Institute of Statistical Mathematics},
  60\penalty0 (4):\penalty0 699--746, Dec 2008.
\newblock ISSN 1572-9052.
\newblock \doi{10.1007/s10463-008-0197-x}.

\bibitem[Sugiyama et~al.(2012)Sugiyama, Suzuki, and
  Kanamori]{sugiyama_suzuki_kanamori_2012}
Sugiyama, Masashi, Suzuki, Taiji, and Kanamori, Takafumi.
\newblock \emph{{Density Ratio Estimation in Machine Learning}}.
\newblock Cambridge University Press, 2012.
\newblock \doi{10.1017/CBO9781139035613}.

\bibitem[Tipping \& Bishop(1999)Tipping and Bishop]{Tipping1999}
Tipping, Michael~E. and Bishop, Christopher~M.
\newblock {Probabilistic Principal Component Analysis}.
\newblock \emph{Journal of the Royal Statistical Society. Series B (Statistical
  Methodology)}, 61:\penalty0 611--622, 1999.
\newblock \doi{10.2307/2680726}.

\bibitem[Titsias(2017)]{Titsias2017}
Titsias, Michalis~K.
\newblock {Learning Model Reparametrizations: Implicit Variational Inference by
  Fitting MCMC distributions}.
\newblock aug 2017.

\bibitem[Tran et~al.(2017)Tran, Ranganath, and Blei]{NIPS2017_7136}
Tran, Dustin, Ranganath, Rajesh, and Blei, David.
\newblock {Hierarchical Implicit Models and Likelihood-Free Variational
  Inference}.
\newblock In Guyon, I, Luxburg, U~V, Bengio, S, Wallach, H, Fergus, R,
  Vishwanathan, S, and Garnett, R (eds.), \emph{Advances in Neural Information
  Processing Systems 30}, pp.\  5527--5537. Curran Associates, Inc., 2017.

\bibitem[Uehara et~al.(2016)Uehara, Sato, Suzuki, Nakayama, and
  Matsuo]{Uehara2016}
Uehara, Masatoshi, Sato, Issei, Suzuki, Masahiro, Nakayama, Kotaro, and Matsuo,
  Yutaka.
\newblock {Generative Adversarial Nets from a Density Ratio Estimation
  Perspective}.
\newblock oct 2016.

\bibitem[Wainwright \& Jordan(2008)Wainwright and Jordan]{Wainwright2008}
Wainwright, Martin~J. and Jordan, Michael~I.
\newblock {Graphical Models, Exponential Families, and Variational Inference}.
\newblock \emph{Foundations and Trends in Machine Learning}, 1\penalty0
  (1–2):\penalty0 1--305, nov 2008.
\newblock ISSN 1935-8237.
\newblock \doi{10.1561/2200000001}.

\bibitem[Zhu et~al.(2017)Zhu, Park, Isola, and Efros]{Zhu2017}
Zhu, Jun-Yan, Park, Taesung, Isola, Phillip, and Efros, Alexei~A.
\newblock {Unpaired Image-to-Image Translation using Cycle-Consistent
  Adversarial Networks}.
\newblock pp.\  2223--2232, mar 2017.

\end{thebibliography}
